\documentclass[twocolumn]{article}

\usepackage{style}

\usepackage[utf8]{inputenc} % allow utf-8 input
\usepackage[T1]{fontenc}    % use 8-bit T1 fonts
\usepackage[square,numbers,comma,sort]{natbib}
\usepackage{amsthm,amsfonts,amsmath,amssymb}       % blackboard math symbols
\usepackage{nicefrac}       % compact symbols for 1/2, etc.
\usepackage{microtype}      % microtypography
\usepackage[usenames,dvipsnames]{color,xcolor}         % colors
\usepackage{lineno}
\usepackage[labelfont=bf,font=bf]{caption}
\usepackage[labelfont=bf,font=bf]{subcaption}
\usepackage{multirow,tabularx,longtable,booktabs}
\usepackage{setspace}
\usepackage{mathrsfs}
\usepackage{pdfpages}
\usepackage{float}
\usepackage{algorithm}
\usepackage{algorithmic}
\usepackage{enumitem}
\usepackage{url}            % simple URL typesetting
\usepackage{booktabs}       % professional-quality tables
\usepackage{hyperref}
\usepackage{authblk}

\hypersetup{colorlinks=true,citecolor=black,filecolor=black,linkcolor=black,urlcolor=black,menucolor=black,
            pdftitle={T-Explainer: A Model-Agnostic Explainability Framework Based on Gradients},
            pdfsubject={Explainable Artificial Intelligence},
            pdfauthor={Ortigossa, et al.},
            pdfkeywords={},
}

\title{T-Explainer: A Model-Agnostic Explainability Framework Based on Gradients}

\author[1]{Evandro S. Ortigossa}
\author[2]{Fábio F. Dias}
\author[3]{Brian Barr}
\author[2]{Claudio T. Silva}
\author[1]{Luis Gustavo Nonato}
\affil[1]{{University of S{\~a}o Paulo (ICMC-USP)}, {S{\~a}o Carlos 13566-590}, {Brazil}}
\affil[2]{{New York University}, {Brooklyn 11201}, {USA}}
\affil[3]{{Capital One}, {McLean 22102}, {USA}}
\affil[ ]{\texttt{evortigosa@usp.br, ffd2011@nyu.edu, brian.barr@capitalone.com, csilva@nyu.edu, gnonato@icmc.usp.br}}

\newcommand{\AI}{Artificial Intelligence}
\newcommand{\ML}{Machine Learning}

\newcommand{\TEXP}{T-Explainer}

\begin{document}

\maketitle

\begin{abstract}
    The development of machine learning applications has increased significantly in recent years, motivated by the remarkable ability of learning-powered systems to discover and generalize intricate patterns hidden in massive datasets. Modern learning models, while powerful, often exhibit a complexity level that renders them opaque black boxes, lacking transparency and hindering our understanding of their decision-making processes. Opacity challenges the practical application of machine learning, especially in critical domains requiring informed decisions. Explainable Artificial Intelligence (XAI) addresses that challenge, unraveling the complexity of black boxes by providing explanations. Feature attribution/importance XAI stands out for its ability to delineate the significance of input features in predictions. However, most attribution methods have limitations, such as instability, when divergent explanations result from similar or the same instance. This work introduces T-Explainer, a novel additive attribution explainer based on the Taylor expansion that offers desirable properties such as local accuracy and consistency. We demonstrate T-Explainer's effectiveness and stability over multiple runs in quantitative benchmark experiments against well-known attribution methods. Additionally, we provide several tools to evaluate and visualize explanations, turning T-Explainer into a comprehensive XAI framework.\vspace{1pt}

\end{abstract}

%=============================================================================
\section{Introduction}
\label{intro}
    Artificial Intelligence is not a vision for the future. It is our present reality.  
    Terms such as Neural Networks, Machine Learning, Deep Learning, and other facets of the Artificial Intelligence universe have seamlessly shifted from futuristic concepts to near-ubiquitous elements in our daily discourse.
    The confluence of recent advances in hardware processing capabilities, abundant data accessibility, and refined optimization algorithms has facilitated the creation of intricate and non-linear machine learning models~\cite{borisov2022deep}. In the contemporary landscape, these models have achieved unprecedented performance levels, surpassing what was deemed inconceivable just a few years ago, outperforming human abilities and previously known methods in central research areas~\cite{ullman2019,snider2022image}.

    The complex non-linear structures and vast number of parameters inherent in such models pose a challenge to transparent interpretation and comprehension of the rationale behind their decisions. This characteristic transforms those models into black boxes in which one can discern only what inputs are provided and what outputs are produced without a clear understanding of the internal decision-making processes~\cite{breiman2001random}. 
    Lack of transparency causes trust-related apprehensions, hindering the effective deployment of more powerful machine learning models in critical applications~\cite{adadi2018xai,arrieta2020,ortigossa2024explainable}. Furthermore, it poses challenges in adhering to emerging regulatory norms observed in numerous countries~\cite{UE2016,ccpa2021}, thereby complicating compliance efforts.
    
    Widely used machine learning models are impenetrable as far as simple interpretations of their mechanisms go~\cite{breiman2001random}.
    In this context, Explainable Artificial Intelligence (XAI) has emerged to address the challenges outlined above, seeking to provide human-understandable insights into the complexities of inherently challenging models. Significantly, the field of explainability has progressed with the introduction of innovative methodologies and research has focused on discerning the strengths and limitations of XAI models~\cite{adadi2018xai}.
    
    Feature attribution/importance methods are particularly relevant, as they are currently the most common explanation type~\cite{tan2023considerations,nauta2023from}. Such methods aim to quantify the contribution of individual input variables to predictions made by black-box models, providing local insights about model reasoning~\cite{ortigossa2024explainable}. 
    While feature attribution methods prove valuable, they have drawbacks that diminish trust and confidence in their application.
    For example, different methods can result in distinct explanations for the same data instances, making it difficult to decide which outcome to believe~\cite{tan2023considerations}. 
    Moreover, many well-known methods, including SHAP~\cite{shap_lundberg2017unified} and LIME~\cite{lime_ribeiro2016should} suffer instability, producing different explanations from different runs on a fixed machine learning model and dataset~\cite{dai2022fairness}.

    An explanation method must produce consistent explanations in multiple runs on the same and similar instances to be considered stable. 
    Stability is a fundamental objective in XAI, as explainability methods that generate inconsistent explanations for similar instances (or even for the same instance) are challenging to trust. 
    If explanations lack reliability, they are essentially useless. 
    Therefore, to be considered reliable, an explanation model must, at a minimum, exhibit stability~\cite{amparore2021}.

    In this work, we introduce {\TEXP}, a novel post-hoc explanation method that relies on the solid mathematical foundations of Taylor expansion to perform local and model-agnostic feature importance attributions.
    {\TEXP} is a local additive explanation technique with desirable properties such as local accuracy, missingness, and consistency~\cite{shap_lundberg2017unified}. 
    Furthermore, {\TEXP} is deterministic, guaranteeing stable results across multiple runs and providing consistent explanations for similar instances.
    
    We conducted several quantitative comparisons against well-known feature attribution methods using controlled synthetic and real-world datasets to evaluate the quality and usefulness of the T-Explainer explanations. Our evaluations focused on consistency, continuity, and correctness of the content explanation~\cite{nauta2023from}. 
    %Since T-Explainer is an additive method\cite{shap_lundberg2017unified} such as SHAP, we also performed an additivity preservation comparison. 
    As a result of such an extensive evaluation process, we implemented a suite of quantitative metrics to evaluate different dimensions of feature importance explainers. These metrics were incorporated into the T-Explainer package, making it not just another XAI method but a comprehensive XAI framework.
    
    In summary, the main contributions of this work are as follows.
    
    \begin{itemize}
        \item \textbf{{\TEXP}}, a stable model-agnostic local additive attribution method. %derived from the solid mathematical foundation of Taylor expansions. 
        It faithfully approximates the local behavior of black-box models using a deterministic optimization procedure, enabling reliable and trustworthy interpretations.
        
        \item A comprehensive set of comparisons against well-known local feature attribution methods using multiple quantitative \textbf{metrics}, aiming to evaluate stability and faithfulness.
        
        \item A Python \textbf{framework} that integrates {\TEXP} with other explainability tools, making the proposed method easy to use and assess in different applications. The T-Explainer, evaluation metrics, datasets, and all related materials are available online.\footnote{%
        %A GitHub link will be provided upon acceptance.
        Available at {\href{https://github.com/evortigosa/EXplainable_AI}{https://github.com/evortigosa/EXplainable\_AI}}
        }
    \end{itemize}

%=============================================================================
\section{Related work}
\label{related}
    Several XAI techniques have been proposed to deal with black-box models, either locally or globally~\cite{tan2023considerations}.
    To contextualize our contributions, we focus the following discussion on post-hoc feature importance/attribution methods~\cite{wojtas2020feature,krishna2022disagreement}. 
    A more comprehensive discussion about XAI can be found in several surveys summarizing existing approaches and their properties~\cite{adadi2018xai,arrieta2020,linardatos2020explainable,ortigossa2024explainable} and metrics to evaluate explanation methods~\cite{yang2019benchmarking,agarwal2022openxai,nauta2023from}.
    
    \citet{breiman2001random} proposed one of the first approaches to identify the features that most impact the predictions of a model. Breiman's solution is model-specific (Random Forests) and involves permuting the values of each feature and computing the model loss. Given the feature independence assumption, the method identifies the most important features by prioritizing those that contribute the most to the overall loss.
     
    More general approaches, the so-called model-agnostic techniques, can (theoretically) operate with any machine learning model, regardless of the underlying algorithm or architecture.
    In this context, LIME (Local Interpretable Model-agnostic Explanations)~\cite{lime_ribeiro2016should}, SHAP (SHapley Additive exPlanations)~\cite{shap_lundberg2017unified}, and their variants~\cite{ortigossa2024explainable} %\cite{lime_ribeiro2016nothing,lime_ribeiro2018anchors,covert2021improving} 
    are model-agnostic methods widely employed to explain machine learning models' behavior in healthcare~\cite{duell2021comparison}, financial market~\cite{gramegna2021shap}, and engineering~\cite{kuzlu2020gaining} applications.
    
    Although widely used, LIME and SHAP have significant drawbacks that require care when using such techniques. 
    For example, LIME lacks theoretical guarantees about generating accurate simplified approximations for complex models~\cite{adadi2018xai}. 
    Moreover, different simplified models can be fitted depending on the random sampling mechanism used by LIME, which can lead to instability to small data perturbation and, sometimes, entirely different explanations by just running the code multiple times~\cite{amparore2021}.

    SHAP has a solid theoretical foundation derived from game theory that grants SHAP desirable properties such as local accuracy, missingness, and consistency~\cite{shap_lundberg2017unified}.
    However, the exact computation of SHAP values is NP-hard, demanding Monte Carlo sampling-based approximations~\cite{tan2023considerations,ortigossa2024explainable}, which introduces instability similar to the LIME method, even in model-specific variants~\cite{shap_lundberg2018consistent,ortigossa2024explainable}. 
    To avoid instability, deterministic versions of LIME~\cite{zafar2021deterministic}, optimization~\cite{lakkaraju2020robust}, and learning-based~\cite{situ2021learning} approaches have been proposed, but at the price of increasing the number of parameters to be tuned.
    
    Gradient-based methods are another important family of explanation approaches.
    Those methods attribute importance to each input feature by analyzing how their changes affect the model's output, relying on gradient decomposition to quantify those effects.
    Vanilla Gradient~\cite{simonyan2013deep} introduced the use of gradients in feature attribution tasks. The method computes partial derivatives at the input point $\mathbf{x}$ with a Taylor expansion around a different point $\mathbf{x}^{\prime}$ and a remainder bias term, for which neither is defined~\cite{bach2015pixel}. 
    Monte Carlo sampling can be used to estimate derivatives, but it makes the Vanilla Gradient suffer from instability and noise within the gradients.

    LRP (Layer-wise Relevance Propagation)~\cite{bach2015pixel} identifies properties related to the maximum uncertainty state of predictions by redistributing an importance score back to the model's input layer. %\cite{lapuschkin2019unmasking,hamilton2021model}
    The Integrated Gradients method~\cite{sundararajan2017axiomatic} quantifies the feature importance by integrating gradients from the target input to a baseline instance. 
    Input $\times$ Gradient highlights influential regions in the input space by computing the features' element-wise product and corresponding gradients from the model's output~\cite{ortigossa2024explainable}. 
    DeepLIFT (Deep Learning Important FeaTures)~\cite{shrikumar2017learning} is based on LRP's importance scores to measure the difference between the model's prediction of a target input and baseline instance.
    
    The stability of gradient-based methods depends on gradient propagation, the model's complexity, and the baseline choice~\cite{shrikumar2017learning}. 
    Additionally, most of those methods are designed specifically for explanation tasks in Neural Networks and other models with differentiable parameters. %, which impairs their application to classifiers such as Random Forests and SVMs.
    
    {\TEXP} differs from the methods described above in three main aspects: 
    (\textit{i}) it is deterministic, {\TEXP} defines an optimization procedure based on finite differences to estimate partial derivatives, thus being stable by definition; 
    (\textit{ii}) {\TEXP} relies on just a few hyperparameters, rendering it easy to use; 
    and (\textit{iii}) {\TEXP} is not dependent on baselines. 
    Moreover, {\TEXP} is built on the solid mathematical foundations of the Taylor expansion, which naturally endows it with desirable properties similar to those of SHAP. 
    Although {\TEXP}, in theory, is designed to be applied to differentiable models, we show experimentally that it also produces interesting results operating with non-differentiable models, which makes {\TEXP} more flexible than previous gradient-based methods.

%=============================================================================
\section{The T-Explainer}
\label{t_exp}
    In this Section, we introduce the theoretical foundations, properties, and computational aspects of the {\TEXP} method.
    
    Let $\mathbf{X}$ be a multidimensional dataset where each data instance $\mathbf{x} = (x_1,\dots, x_n) \in \mathbf{X}$ is a vector in $\mathbb{R}^n$ and $f$ be a \MakeLowercase{\ML} model. For simplicity, let's assume that $f$ is a binary classification model trained on $\mathbf{X}$, that is, $f(\mathbf{x}) \in (0,1)$ accounts for the probability of $\mathbf{x}$ belonging to class $1$, and $(1 - f(\mathbf{x}))$ is the probability of belonging to class $0$. All the following reasoning can be extended to regression models. 
    
    The model $f$ can be seen as a real-valued function:
    
    \begin{equation}
        \label{eq:f}
        f : \mathbf{X} \rightarrow (0,1) \subset {\rm I\!R}
    \end{equation}
    
    \noindent As a real function, $f$ can be linearly approximated through first-order Taylor expansion:
    
    \begin{equation}
        \label{eq:taylor}
        f(\mathbf{x} + \mathbf{h}) \approx f(\mathbf{x}) + \nabla f(\mathbf{x}) \cdot \mathbf{h}
    \end{equation}
    
    \noindent where $\mathbf{h}$ is a displacement vector corresponding to a small neighborhood perturbation of $\mathbf{x}$ and $\nabla f(\mathbf{x})$ is the gradient (linear transformation) of $f$ in $\mathbf{x}$, given by:
    
    \begin{equation}
        \label{eq:grad}
        {\nabla}f(\mathbf{x}) = \left[{\frac{\partial f(\mathbf{x})}{\partial x_1}}, \dots , {\frac{\partial f(\mathbf{x})}{\partial x_n}}\right].
    \end{equation}

    The $i$-th gradient element corresponds to the partial derivative of $f$ concerning the $i$-th attribute of $\mathbf{x}$. 
    Note that the gradient of $f$ in $\mathbf{x}$ corresponds to the Jacobian matrix when $f$ is a real-valued function. 
    The dot product between the gradient and the displacement vector $\mathbf{h}$ is a linear map from $\mathbb{R}^n$ to $\mathbb{R}$, well known as the best linear approximation of $f$ in a small neighborhood of $\mathbf{x}$.%\cite{press2007numerical}
    Therefore, the gradient can be used to analyze how small perturbations in the input data affect the model output.
    The right side of Equation~\ref{eq:taylor} is a linear equation that approximates the behavior of $f$ near $\mathbf{x}$, and because it is a linear mapping, it is naturally interpretable. 
    In other words, the gradient of a model $f$ can be used to generate explanations.

    The formulation above resembles the Vanilla Gradient~\cite{simonyan2013deep} Taylor expansion-based procedure to compute saliency maps. The difference is that, in our case, the attributions are not dependent on class information and do not rely on further parameters specific to the model's architecture. In addition, T-Explainer differs from previous gradient-based methods by relying on additive modeling and a deterministic optimization procedure to approximate gradients, as detailed below.

    Let $\mathbf{h}=\mathbf{z}^{\prime} \in \mathbb{R}^n$ be a perturbation in $\mathbf{x}$, that is, $\mathbf{x}^{\prime}=\mathbf{x}+\mathbf{z}^{\prime}$ is a point in a small neighborhood of $\mathbf{x}$. 
    The {\TEXP} is defined as an additive explanation modeling $g_{\mathbf{x}}$, given by:
    
    \begin{equation}
        \label{eq:add_ft_att}
        \setlength{\abovedisplayshortskip}{-3pt}
        g_{\mathbf{x}}(\mathbf{z}^{\prime}) = \phi_0 + \sum_{i=1}^{n}{\phi_i {z}^{\prime}_i}
    \end{equation}
    
    \noindent where $\phi_0=\mathrm{\mathbb{E}}[f(\mathbf{X})]$ represents the expected prediction value, and $\phi_i = \frac{\partial {f}(\mathbf{x})}{\partial x_i}$.
    The prediction expected value is a statistical value that is not trivial to estimate for an arbitrary dataset.
    In practice, it is computed through the average model output across the training dataset $\mathbf{X}$ when the feature values $\mathbf{X}_i$ are unknown.
    Specifically, it does not mean that $\mathbf{X}_i = 0$; it means we do not know the value of $\mathbf{X}_i$. Therefore, its distribution is estimated based on the data by taking the mean value and then averaging the predicted probabilities for each label.
    As a fundamental property of additive explanations, Equation~\ref{eq:add_ft_att} approximates the original predicted value $f(\mathbf{x})$ locally by summing its feature importances~\cite{shap_lundberg2017unified}.
    
    The explanation model $g_{\mathbf{x}}$ is a local attribution method, i.e., there is a $g_{\mathbf{x}}$ for each $\mathbf{x}$. 
    By definition, {\TEXP} is an \textit{additive feature attribution method}, as defined by \citet{shap_lundberg2017unified}, which means that the importance value attributed to each feature can reconstruct the model prediction by adding those importance values. 
    The coefficient $\phi_i$ indicates the feature attribution/importance of the $i$-th attribute to the prediction made by $f$ in $\mathbf{x}$. 
    In {\TEXP}, the feature attribution $\phi_i$ has a simple and intuitive geometric interpretation, corresponding to the projection of $\nabla f(\mathbf{x})$ on the $i$-th feature axis. 
    Therefore, the more aligned the gradient $\nabla f(\mathbf{x})$ and the $i$-th axis of the feature space, the more important the $i$-th feature is to the decision made by $f$.

%-----------------------------------------------------------------------------
\subsection{T-Explainer Properties}
\label{t_exp:props}
    According to \citet{shap_lundberg2017unified}, a ``good'' explanation method must have three important properties, namely, \textit{Local Accuracy}, \textit{Missingness}, and \textit{Consistency}. In the following, we show that {\TEXP} approximates \textit{Local Accuracy} while holding \textit{Missingness} and \textit{Consistency}.

%-----------------------------------------------------------------------------
    \subsubsection{Local Accuracy}
    \label{t_exp:props:local}
        The local accuracy property, as defined by \citet{shap_lundberg2017unified}, states that if $f(\mathbf{x} + \mathbf{z}^{\prime}) = g_{\mathbf{x}}(\mathbf{z}^{\prime}) = \phi_0 + \sum_{i=1}^{n}{\phi_i {z}^{\prime}_i}$, then $g_{\mathbf{x}}$ has the local accuracy property. The {\TEXP} does not fully satisfy this property but instead approximates it. By construction, we have the following:
        
        \begin{equation}
            \label{eq:local}
            \setlength{\abovedisplayshortskip}{-6pt}
            f(\mathbf{x} + \mathbf{z}^{\prime}) \approx g_{\mathbf{x}}(\mathbf{z}^{\prime}) = \phi_0 + \sum_{i=1}^{n}{\phi_i {z}^{\prime}_i}
        \end{equation}
        
        \noindent and, from the Taylor expansion remainder theorem~\cite{marsden2003vector}, there is an upper bound to the approximation error given by:
        
        \begin{equation}
            \label{eq:local_error}
            {f(\mathbf{x} + \mathbf{z}^{\prime}) - g_{\mathbf{x}}(\mathbf{z}^{\prime}) = O({\| \mathbf{z}^{\prime} \|}^2)}.
        \end{equation}

%-----------------------------------------------------------------------------
    \subsubsection{Missingness}
    \label{t_exp:props:miss}
        The missingness property states that if a feature $x_i$ has no impact on the model decision, then $\phi_i = 0$~\cite{shap_lundberg2017unified}. 
        In our context, a feature $i$ having no impact on $f$ means that $f$ does not vary (increase or decrease) when only the $i$-th feature is changed (otherwise, the feature would impact the model decision). 
        In other words, $f(x_1,\dots, x_i + {z}^{\prime}_i,\dots, x_n) - f(x_1,\dots, x_i,\dots, x_n) = 0$, thus, there is no variation in the $i$-th direction and the partial derivative $\phi_i = \frac{\partial f(\mathbf{x})}{\partial x_i} = 0$, ensuring that the {\TEXP} holds the missingness property.

%-----------------------------------------------------------------------------
    \subsubsection{Consistency}
    \label{t_exp:props:consist}
        Let $f$ and $\tilde{f}$ be two binary classification models. Let us use the notation $\mathbf{x}^{\prime} \backslash i$ to indicate that the $i$-th feature is disregarded in any perturbation of $\mathbf{x}$ (${z}^{\prime}_i = 0$ in any perturbation, so ${x}^{\prime}_i \backslash i = x_i$). An explanation method is consistent if (see \citet{shap_lundberg2017unified}), fixing $\mathbf{x}$, $\tilde{f}(\mathbf{x}^{\prime}) - \tilde{f}(\mathbf{x}^{\prime} \backslash i) > f(\mathbf{x}^{\prime}) - f(\mathbf{x}^{\prime} \backslash i)$ implies $\phi_{i}(\tilde{f}) > \phi_{i}(f)$.
        Suppose that $\tilde{f}(\mathbf{x}^{\prime}) - \tilde{f}(\mathbf{x}^{\prime} \backslash i) > f(\mathbf{x}^{\prime}) - f(\mathbf{x}^{\prime} \backslash i)$ holds in a small neighborhood of $\mathbf{x}$, in particular,
        %
        % Display-math environments should never, ever occur at the start of a paragraph.
        \begin{align}
            \label{eq:consist}
            \tilde{f}(x_{1}+{z}^{\prime}_{1}, \dots, x_{i}&+{z}^{\prime}_{i}, \dots, x_{n}+{z}^{\prime}_{n}) - \nonumber\\
            &\tilde{f}(x_{1}+{z}^{\prime}_{1}, \dots, x_{i}, \dots, x_{n}+{z}^{\prime}_{n}) > \\
            f(x_{1}+{z}^{\prime}_{1}, \dots, & x_{i}+{z}^{\prime}_{i}, \dots, x_{n}+{z}^{\prime}_{n}) - \nonumber\\
            &f(x_{1}+{z}^{\prime}_{1}, \dots, x_{i}, \dots, x_{n}+{z}^{\prime}_{n})\nonumber
        \end{align}
        
        \noindent for ${z}^{\prime}_{i} \in (-\delta, 0) \cup (0, \delta)$.
        
        Define $\tilde{s}({z}^{\prime}_{i}) = \frac{\tilde{f}(\mathbf{x}^{\prime}) - \tilde{f}(\mathbf{x}^{\prime} \backslash i)}{{z}^{\prime}_{i}}$ and ${s}({z}^{\prime}_{i}) = \frac{{f}(\mathbf{x}^{\prime}) - {f}(\mathbf{x}^{\prime} \backslash i)}{{z}^{\prime}_{i}}$, from Equation~\ref{eq:consist} we know that $\tilde{s}({z}^{\prime}_{i}) > {s}({z}^{\prime}_{i})$ for ${z}^{\prime}_{i} \in (-\delta, 0) \cup (0, \delta)$. Assuming that $\tilde{f}$ and ${f}$ are differentiable in $\mathbf{x}$, then $\lim_{{z}^{\prime}_{i}\to 0}\tilde{s}({z}^{\prime}_{i})=\frac{\partial \tilde{f}(\mathbf{x})}{\partial x_i}$ and $\lim_{{z}^{\prime}_{i}\to 0}{s}({z}^{\prime}_{i})=\frac{\partial {f}(\mathbf{x})}{\partial x_i}$ exist, thus, from the \textit{Limit Inequality Theorem}:
        
        \begin{equation}
            \label{eq:lim_ineq_theo}
            \phi_{i}(\tilde{f})=\frac{\partial \tilde{f}(\mathbf{x})}{\partial x_i} > \frac{\partial {f}(\mathbf{x})}{\partial x_i}=\phi_{i}({f})
        \end{equation}
        
        \noindent showing that the {\TEXP} holds the consistency property (the \textit{Limit Inequality Theorem} ensures that, given two functions $\tilde{s}, s : (a,c)\cup(c,b) \subset {\rm I\!R} \rightarrow {\rm I\!R}$, if $\tilde{s}(x) > s(x)$ for all $x \in (a,c)\cup(c,b)$ and the limits $\lim_{x\to c}\tilde{s}(x)=A$ and $\lim_{x\to c}{s}(x)=B$ exist, then $A > B$).
        
        According to \citet{shap_lundberg2017unified}, the Shapley-based explanation is the unique possible additive feature attribution model that (theoretically, see \citet{hooker2021unrestricted}) satisfies \textit{Local Accuracy}, \textit{Missingness}, and \textit{Consistency} properties. As demonstrated above, {\TEXP} approximates \textit{Local Accuracy} while holding \textit{Missingness} and \textit{Consistency}. Therefore, the {\TEXP} is one of the few XAI methods that approximate SHAP theoretical guarantees.

%-----------------------------------------------------------------------------
    \subsection{T-Explainer: Computational Aspects}
    \label{t_exp:development}
        \begin{figure*}[!t]
            \centering
            \includegraphics[width=0.9\textwidth]{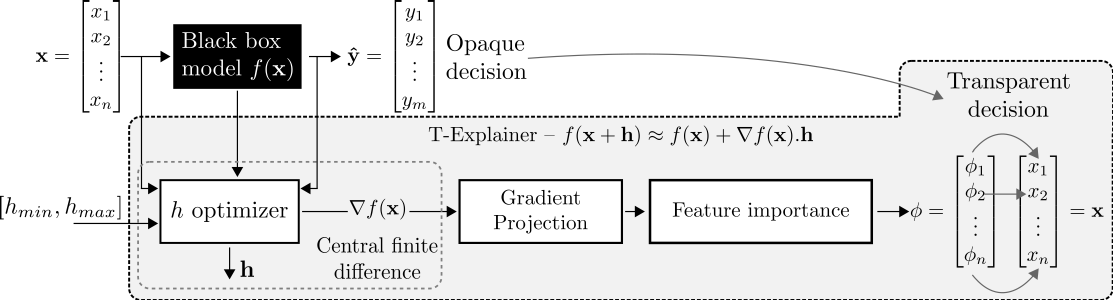}
            \caption{{\TEXP} pipeline.}
            \label{fig:t_exp_pipeline}
        \end{figure*}
            
        Figure~\ref{fig:t_exp_pipeline} illustrates the step-by-step pipeline of {\TEXP} for feature attribution.
        Computing the gradient of a known real-valued function is (theoretically) simple, demanding that the partial derivatives of the function be taken. However, we need to calculate the partial derivatives of an arbitrary black-box model $f$, which was previously trained and has complex internal mechanisms. To that end, we perturb the instance $\mathbf{x}$ attribute-wise and compute the respective change in the model's output, approximating the partial derivatives through centered finite differences~\cite{leveque2007finite}:
        
        \begin{equation}
            \label{eq:center_findif}
            FD_{f(\mathbf{x})}=\frac{f(\mathbf{x}+\mathbf{h})-f(\mathbf{x}-\mathbf{h})}{2||\mathbf{h}||}.
        \end{equation}
        
        Finite differences are well-established approaches to approximate differential equations, replacing the derivatives with discrete approximations (such transformation results in computationally feasible systems of equations)~\cite{leveque2007finite}.
        Specifically, we rely on the centered finite difference because that approach averages two one-sided perturbations of each attribute, resulting in a second-order accurate approximation with an error proportional to $||\mathbf{h}||^2$~\cite{leveque2007finite}. 
        The centered finite difference approximation of the partial derivatives demands establishing the magnitude of the perturbation $\mathbf{h}$. 
        The displacement $\mathbf{h}$ must be small enough to generate a perturbation close to the input data $\mathbf{x}$. 
        By definition, the derivatives of $f$ are computed by making $\mathbf{h}\to 0$ in Equation~\ref{eq:center_findif}.
        In other words, we have to set $\mathbf{h}$ to a small value.
        
        In practice, if $\mathbf{h}$ is too small, it can generate significant round-off errors or singularity cases. 
        However, if $\mathbf{h}$ is too large, it can lead to truncation errors and inappropriate approximations. 
        To our knowledge, there is no closed solution to determine an optimum value of $\mathbf{h}$.
        To solve this issue, we developed a $\mathbf{h}$ optimizer method based on a binary search that minimizes a \emph{mean squared error} (MSE) cost function.
        The method runs over the $[h_{min}, h_{max}]$ interval, searching for an optimal $\mathbf{h}$ to produce a good approximation of $f(\mathbf{x})$ from Equation~\ref{eq:taylor}. 
        We set $h_{min}$ with the minimum distance between any two instances on the dataset, i.e., $h_{min}=\min_{i,j=1}^{n}||{\mathbf{x}_i}-{\mathbf{x}_j}||$, $\forall {i}\neq{j}, \mathbf{x}_i,\, \mathbf{x}_j \in \mathbf{X}$.
        The optimization method is formulated as follows.
        
        \begin{equation}
            \label{eq:optimize_h}
            \nabla f(\mathbf{x}) = \underset{\mathbf{h} \in [h_{min}, h_{max}]}{\mathrm{arg\,min}}\ \mathcal{L}(f,\mathbf{h},\epsilon_{\mathbf{x}}) + \theta(\nabla f_{\mathbf{x}})
        \end{equation}
        
        \noindent where $\mathcal{L}$ is the cost function (MSE in our case),
        $\nabla f_{\mathbf{x}}$ represents the approximated value of the gradient $\nabla f(\mathbf{x})$ in each iteration of the $\mathbf{h}$ optimizer algorithm computing the method in Equation~\ref{eq:center_findif}
%        on Equation~\ref{eq:taylor}
        , $\epsilon_{\mathbf{x}}$ is the upper limit of the cost threshold bound by $(h_{min})^2$~\cite{leveque2007finite}, and $\theta$ is a method that ensures the numerical stability of the optimization process by checking if $\nabla f(\mathbf{x})$ comes from a non-singular (or full-rank) Jacobian matrix. 
        
        In the presence of a singular matrix, perturbation methods such as finite differences suffer from numerical instability. This instability can lead to inaccurate approximations of the Jacobian, making it unreliable for further calculations or updates. In addition, a singular or rank-deficient Jacobian indicates that some input features or dimensions are linearly dependent, and this is problematic because it means that the model is not fully sensitive to certain input variations, leading to poor sensitivity analysis. Specific to the context of our optimizer, a singular Jacobian leads to a failure in convergence, propagating errors that lead the explainer to output one-hot-like feature importance vectors. Therefore, $\theta$ rejects $\mathbf{h}$ when it results in a singular (or rank-deficient) Jacobian.

        It is important to highlight that local perturbations must preserve the normalization range of the data to avoid model extrapolations~\cite{hooker2021unrestricted}.
        Further details about the optimization procedure are given in the Appendix.
        This approach can find the gradient ${\nabla f(\mathbf{x})}$ of a binary-class model or even the entire Jacobian related to all the model's output probabilities. 
        Once the gradient is estimated at $\mathbf{x}$, the importance of each attribute is computed as $\phi_i = \frac{\partial {f}(\mathbf{x})}{\partial x_i}$.
        
        %Equation~\ref{eq:taylor} represents the best linear approximation of a function $f$ for perturbed points close to $\mathbf{x}$, describing how input perturbations locally influence the model's output. 
        The explanations must remain consistent with the model's behavior in the instance's locality to ensure meaningful interpretability.
        Through the $\mathbf{h}$ optimization process, T-Explainer attributes feature importance using Taylor's expansion for each instance tailored to its local characteristics.  
        In other words, the T-Explainer is designed primarily as a local explainer.
        According to \citet{lime_ribeiro2016should}, simultaneously achieving local and global fidelity in XAI is challenging because the aggregation of local explanations to estimate global feature importance could be ineffective. After all, local explanations are instance-specific and often inconsistent with global explanations.
        However, the adaptive $\mathbf{h}$ optimization strategy of {\TEXP} is flexible enough to be extended and generate aggregated views while preserving local relationships, enhancing consistent and interpretable explanations at local and global levels.

%-----------------------------------------------------------------------------
    \subsubsection{Handling Categorical Data}
    \label{t_exp:development:cat_data}
        All the considerations outlined in the T-Explainer's theoretical foundations (see The T-Explainer Section) assume that the attributes are continuous numerical data. However, most datasets are not limited to numerical data but also encompass categorical (nominal or qualitative) features. Handling categorical data is challenging, as many learning models cannot directly process nominal features, demanding numerical conversions. Different approaches can be applied to encode categorical values as numerical representations, including the well-known one-hot encoding~\cite{RODRIGUEZ201821}.
    
        Numerical encodings introduce challenges for XAI methods, particularly those based on gradients. 
        Nominal values represented with one-hot encode become binary constant values (zeros and ones), thus being discontinuous attributes where partial derivatives can not be properly estimated, inducing the explainer to erroneously attribute null importance to features that may significantly impact the prediction.
        
        T-Explainer has a modular design that enhances flexible additions or improvements in functionalities. 
        To address the challenges of categorical features, we implemented a mechanism to handle one-hot encoded columns into the T-Explainer framework. 
        Once the user identifies the one-hot encoded categorical attributes, a procedure transforms them into intervals through continuous perturbations.
        Specifically, the values $0$ and $1$ (resulting from one-hot encoding) are uniformly perturbed with displacement in the interval $\delta \in (-0.5, 0.5)$, creating a range of values around $0$ and $1$, simulating a continuous set of values around these two values. 
        
        The continuity induction procedure is performed on a copy of the training dataset, with the perturbed columns normalized to the same interval of the numerical features (typically in the $[0,1]$ range). After that, transfer learning is applied to a copy of the original model, fitting it with the perturbed training dataset (the categorical fitted model). Finally, the T-Explainer runs on the newly trained model.
        In summary, predictions of instances holding categorical features are explained using the T-Explainer's core implementation. However, we incorporated a preliminary layer in T-Explainer's pipeline to handle categorical data.
    
        The perturbation radius $\delta$ can affect the accuracy of the categorical fitted model, making it disagree with the original model. 
        However, the disagreement varies according to the training data.
        In our experiments, the proposed transformation of categorical attributes to numerical ones does not significantly impact the accuracy of the categorical fitted model for perturbations in small intervals.
        We fixed $\delta= 0.1$ as the perturbation radius, which ensured accuracy preservation.

%=============================================================================
\section{Quantitative Metrics}
\label{bench}
    A common drawback of research proposing XAI methods is a lack of clarity on the use of quantitative evaluations~\cite{nauta2023from}. Most proposals rely on visual inspections or simplified human-centered case studies to check whether an explanation ``looks reasonable.'' However, such anecdotal strategies are prone to subjectivities and do not allow for formal comparisons between explanation methods, which is insufficient to ensure a robust verification of explanations' consistency, continuity, and correctness~\cite{nauta2023from}. Evaluation in XAI evolved significantly in the past few years, and this Section describes the evaluation metrics integrated into the T-Explainer framework. %We focused our evaluations on automated quantitative metrics that enable us to compare different explanation methods regarding their stability, faithfulness, and local accuracy preservation.
    
    A stability metric measures the robustness or sensitivity of an explainer when exposed to slightly different versions of an original input sample. \textit{Relative Input Stability} (RIS) and \textit{Relative Output Stability} (ROS)~\cite{agarwal2022openxai} are metrics used to evaluate the stability of local explanations to changes in input data and output prediction probabilities, respectively, based on the ``relative stability'' formulation that applies normalized distances allowing direct comparisons between different feature importance methods.
    %\citet{alvarez2018towards} introduced the stability concept, which \citet{agarwal2022rethinking} recently enhanced through the ``relative stability'' formulation. 
    We followed the theoretical definitions of RIS and ROS defined by \citet{agarwal2022openxai}.
    
    Given $\mathcal{N}_{\mathbf{x}}$ a neighborhood of perturbed samples $\mathbf{x}^{\prime}$ around an original input instance $\mathbf{x}$, with $e_{\mathbf{x}}$ and $e_{\mathbf{x}^{\prime}}$ representing the attribution vectors explaining $\mathbf{x}$ and $\mathbf{x}^{\prime}$, and $f({\mathbf{x}})$ and $f({\mathbf{x}^{\prime}})$ the output prediction probabilities of $\mathbf{x}$ and $\mathbf{x}^{\prime}$, RIS and ROS metrics are defined as:
    
    \begin{equation}
        \label{eq:ris}
        \setlength{\abovedisplayshortskip}{-6pt}
        \text{RIS}(\mathbf{x},\mathbf{x}^{\prime},e_{\mathbf{x}},e_{\mathbf{x}^{\prime}})= \max_{\mathbf{x}^{\prime}}{\frac{\| \frac{e_{\mathbf{x}} - e_{\mathbf{x}^{\prime}}}{e_{\mathbf{x}}} \|_{p}}{\max{(\| \frac{\mathbf{x} - \mathbf{x}^{\prime}}{\mathbf{x}}\|_{p}, \epsilon_{c})}}}\:\:\:\:\text{and}
    \end{equation}
    
    \begin{equation}
        \label{eq:ros}
        \setlength{\abovedisplayshortskip}{-6pt}
        \text{ROS}(\mathbf{x},\mathbf{x}^{\prime},e_{\mathbf{x}},e_{\mathbf{x}^{\prime}})= \max_{\mathbf{x}^{\prime}}{\frac{\| \frac{e_{\mathbf{x}} - e_{\mathbf{x}^{\prime}}}{e_{\mathbf{x}}} \|_{p}}{\max{(\| \frac{f(\mathbf{x}) - f(\mathbf{x}^{\prime})}{f(\mathbf{x})}\|_{p}, \epsilon_{c})}}}
    \end{equation}
    
    \noindent $\forall \mathbf{x}^{\prime} \in \mathcal{N}_{\mathbf{x}}$, $\hat{y}_{\mathbf{x}} = \hat{y}_{\mathbf{x}^{\prime}}$, with ${p}$ defining the $l_p$ norm used to measure the changes and $\epsilon_{c}>0$ as a small value threshold to avoid zero division. 
    The higher the RIS or ROS values, the more unstable the method is related to input or output prediction perturbation.
    
    We identified some issues that prevented us from using the implementations of \citet{agarwal2022openxai}. Notably, their framework lacks flexibility for incorporating novel explainers beyond those included in OpenXAI, and the separate execution of RIS and ROS metrics reduces their comparability due to differences in the perturbed samples drawn from normal distributions. In contrast, our T-Explainer framework offers a flexible Python module that can be readily extended to assess multiple feature attribution methods while computing RIS and ROS in parallel.

    Previously, the selection of perturbed samples satisfying $\hat{y}{\mathbf{x}} = \hat{y}{\mathbf{x}^{\prime}}$ was performed randomly, with no guarantee that a sufficient number of samples would be retained -- particularly near the model's decision boundaries. We solved this limitation by sampling perturbations for each $\mathbf{x}$ and retaining that $\hat{y}_{\mathbf{x}} = \hat{y}_{\mathbf{x}^{\prime}}$ according to their distance to $\mathbf{x}$~\cite{lime_ribeiro2016should}, ensuring a neighborhood holding a minimum set of samples.
    
    RIS and ROS handle values close to zero in normalization processes. Both large negative and positive values indicate high importance -- the sign indicates the direction of the feature's contribution to the prediction. Therefore, 
    %$\epsilon_{c}$ cannot be a cutoff that ignores negative values by trimming them as small values. Then, 
    we designed a double-sided clipping method that preserves significant values, either positive or negative, while discarding only those close to zero.
    In addition, we extended RIS and ROS to provide mean stability values. %and the respective standard deviations, in addition to the original maximum change in explanations. 
    %With these improvements, RIS and ROS can assess the stability and indicate whether the model (and consequently the XAI method) handles outliers in the data.

    To complement our stability evaluation, we introduce the \textit{Run Explanation Stability} (RES) metric to quantify the stability of local explainers across multiple runs on non-perturbed inputs. 
    The core idea behind RES is straightforward: Under fixed model and explanation method settings, the same original input should yield identical explanations~\citep{nauta2023from}. In practice, we generate $n$ explanations, $e_{\mathbf{x}1}, \dots, e_{\mathbf{x}n}$, for the same original instance $\mathbf{x}$ and compute their standard deviation, returning the maximum deviation in a set of instances. The RES metric is defined as follows:
    %e introduce the \textit{Run Explanation Stability} (RES) metric to quantify the stability of local explainers for multiple runs on non-perturbed inputs. 
    %The principle behind RES is simple: Generate multiple explanations for the same original input under fixed model and explanation method settings and measure the extent to which such explanations change. According to the \textit{consistency} property, identical inputs have identical explanations.\cite{nauta2023from}
    %Given $n$ explanations $e_{\mathbf{x}i},\dots,e_{\mathbf{x}n}$ of the same non-perturbed instance $\mathbf{x}$, let $\bar{e}_{\mathbf{x}}$ be the mean explanation from those $n$ explanations, RES is defined as:

    \begin{equation}
        \label{eq:res}
        \setlength{\abovedisplayshortskip}{-6pt}
        \text{RES}(e_{\mathbf{x}},\bar{e}_{\mathbf{x}})= \max_{k=1}^{n}{\| \bar{e}_{\mathbf{x}}-e_{\mathbf{x}k} \|}\text{,}\:\:\: \forall \mathbf{x} \in \mathbf{X}\text{.}
    \end{equation}

    Although we introduce RES as part of the T-Explainer framework, the theoretical foundation behind it came from the Reiteration Similarity metric~\cite{amparore2021}; however, we simplified the formulation by using the standard deviation as the similarity measure.
    
    RIS, ROS, and RES metrics are unitless quantifiers with no ``ideal'' desirable values. Reasonable values depend on the context of the application, meaning those metrics must be interpreted relatively by comparing their results across XAI methods~\cite{alvarez2018towards}. 
    However, lower values indicate methods with higher stability rates.
    Therefore, RIS and ROS fit into the \textit{stability for slight variations} strategy of the \textit{continuity} category of XAI assessment, while RES fits in the \textit{consistency} category~\cite{nauta2023from}.

    A faithfulness metric evaluates the extent to which an explanation is faithful to the predictive model it explains. In this sense, we implemented the \textit{Prediction Gap on Important Features} (PGI)~\cite{petsiuk2018rise,dai2022fairness} metric to evaluate the methods' faithfulness under the \textit{single/incremental deletion} strategy for \textit{correctness}~\cite{nauta2023from}. 
    %\citet{petsiuk2018rise} introduced the intuition behind deletions, which \citet{dai2022fairness} recently enhanced in the PGI metric. 
    Given $top(k, e_{\mathbf{x}})$ being the $k$ most important features determined by a local explanation $e_{\mathbf{x}}$, we implemented PGI iterating on $m$ input instances as follows:

    \begin{equation}
        \label{eq:pgi}
        \setlength{\abovedisplayshortskip}{-4pt}
        \text{PGI}(\mathbf{x},f, e_{\mathbf{x}})= \frac{1}{m}\sum_{j=1}^{m} [|f(\mathbf{x}) - f(\mathbf{\tilde{x}}_j)|]
    \end{equation}

    \noindent where ${\tilde{x}_i} = 0 \,\,\,[i \in top(k, e_{\mathbf{x}})]$ and $f$ return predicted probabilities. PGI values will be in the $[0, 1]$ interval; the higher the value, the more faithful the explanations are compared to the ``true'' black-box behavior, as perturbing those $k$ features caused a significant change in prediction.
    
    %We also introduce the \textit{Local Accuracy Preservation} (LAP) metric, which evaluates the extent to which feature importance explainers preserve the local accuracy property, i.e., the model prediction should be reconstructed by the summation of the importance values (see Equation~\ref{eq:local}).
    %Given an output prediction probability $f(\mathbf{x})$ and its respective explanation $g_{\mathbf{x}}$, we define the LAP metric as the frequency ratio in which $| f(\mathbf{x}) - g_{\mathbf{x}} | \leq \epsilon$, $\forall \mathbf{x} \in \mathbf{X}$, with $\epsilon$ representing a tolerance error and $g$ following the additive modeling defined in Equation~\ref{eq:add_ft_att}.\cite{shap_lundberg2017unified} 
    %The tolerance error is necessary since we do not expect any explanation modeling $g_{\mathbf{x}}$ to achieve a perfect match to the original model $f(\mathbf{x})$.
    %LAP returns values in the $[0, 1]$ range; the higher the value, the more faithful the explainer is regarding local accuracy preservation, i.e., the more an explainer keeps $g_{\mathbf{x}}$ close to $f(\mathbf{x})$, $\forall \mathbf{x} \in \mathbf{X}$, the more ``locally accurate'' it is.

    Unlike previous implementations, we developed metrics to run all the explainers under evaluation in the same execution cycles. It is an advantage regarding fair comparisons since it ensures the explainers are exposed to the same conditions. 
    Our evaluations are based on automated quantitative metrics, enabling users to formally compare the performances of state-of-the-art feature importance methods. 
    The selected metrics evaluate the consistency, continuity, and correctness dimensions of XAI evaluation~\cite{nauta2023from}, providing robust experimental results beyond the typical evaluations in the XAI literature.

%=============================================================================
\section{Experiments}
\label{experiments}
    This section presents the configurations and outcomes of the experiments undertaken to evaluate {\TEXP}, employing a diverse set of datasets, models, and comparison metrics. In all experiments described below, we are evaluating local explanation tasks.

%-----------------------------------------------------------------------------
    \subsection{Experimental Setup}
    \label{experiments:setup}
        We trained Neural Networks from \textit{scikit-learn} library\footnote{\href{https://scikit-learn.org/}{https://scikit-learn.org/}} and Gradient-boosted Tree Ensembles %classifiers (Random Forests-based) 
        from the \textit{XGBoost} %gradient boosting 
        library\footnote{\href{https://xgboost.readthedocs.io/en/stable/}{https://xgboost.readthedocs.io/}} as the black-box models used throughout the experiments. 
        Logistic Regressions and SVMs could also be viable alternatives, but we focused on Neural Networks and Random Forests due to their wide adoption in Machine Learning.
        The models were trained with a dataset split of $80\%$ for training and $20\%$ for testing. Such division was achieved using the $train\_test\_split$ method from \textit{scikit-learn}, with a consistent shuffling seed across all datasets. We conducted a Grid Search to determine the hyperparameters that optimize the models' performance.
        
        For the Neural Networks, we initialized the hyperparameters search using powers of $2$ for the number of neurons in the hidden layers, following standard practices in this context~\cite{tan2023considerations}. 
        We employed the ReLU activation function, Stochastic Gradient Descent as the optimizer and the log-loss function.
        We evaluated alternative activation functions and optimizers, but they did not yield gains.
        The Tree Ensemble model's hyperparameters were defined by specifying ranges for the number of decision tree estimators and the maximum depth of each estimator, utilizing cross-entropy loss as the evaluation metric for classification.
        
        We selected three different neural networks from hyperparameter tuning: Three- and five-layer neural networks that hold $64$ neuron units per layer (3H-NN and 5H-64-NN, with 9.7K and 18K trainable parameters, respectively) and a five-layer neural network with $[64, 128, 128, 128, 64]$ neurons in each layer (5H-128-NN) and roughly 51K trainable parameters.
        All neural network models use a learning rate of $0.01$, alpha $0.0001$, and maximum training epochs of $500$.
        The selected neural networks achieved better performances considering our classification tasks.
        We based our architectures on previous works~\cite{baldi2014searching,borisov2022deep}, with extra neural units and additional layers significantly increasing the training time without noticeably increasing performance.
        
        The Tree Ensemble classifiers (XRFC) have $500$ estimators, a maximum depth of $6$, a learning rate of $0.01$, and a gamma equal to $1$. 
        All the other hyperparameters are the libraries' default for Neural Networks and Tree Ensembles. 
        Those models are used as the base black boxes for comparing the T-Explainer against SHAP~\cite{shap_lundberg2017unified}, LIME~\cite{lime_ribeiro2016should}, and three gradient-based methods from the Captum library~\cite{kokhlikyan2020captum} -- Integrated Gradients, Input $\times$ Gradient, and DeepLIFT. 
        Specifically, we used the SHAP explainer for neural networks and the TreeSHAP explainer~\cite{shap_lundberg2018consistent} for tree-based classifiers. 
        Although SHAP, LIME, and Gradient-based methods were proposed a few years back, we benchmark T-Explainer with them because they continue to be the most widely used feature attribution explainers in research and practice~\cite{agarwal2022openxai}.
        
        For each experiment requiring data perturbation, we used the \textit{NormalPerturbation} method from OpenXAI~\cite{agarwal2022openxai} to generate perturbed neighborhoods with $\mu=0$, $\sigma^{2}=0.001$, a flip percentage $\varepsilon_{p}=0.0001$, and the perturbation maximum distance of $h_{min}/2$, ensuring neighborhoods with small perturbations around each instance. We also specified the clipping threshold to handle values close to zero as $\epsilon_{c}=\pm{10^{-5}}$ and the $l_p$ norm as the Euclidean ($p=2$). 

        To improve readability and avoid roundoff issues, we use scientific notation to represent values larger than ${10}^{5}$, while values smaller than ${10}^{-10}$ will be taken as zero.

%-----------------------------------------------------------------------------
    \subsection{Synthetic Data}
    \label{experiments:synth}
        We generate two different synthetic datasets comprising $1,000$ instances each. The first is a 4-dimensional (4-FT) dataset where each instance $\mathbf{x}$ is generated as follows. We distribute the target label $\mathbf{y} \in [0, 1]$ equally across each dataset half; thus, each class has $500$ instances. Conditioned on the value of $\mathbf{y}$, we sample the instance $\mathbf{x}$ as $x_{1:2} \sim \mathcal{N}(\mu_{y},\,\Sigma_{y})$. We choose $\mu_0=[0,0]^{\mathrm {T}}$ and $\mu_1=[-2,2]^{\mathrm {T}}$, $\Sigma_0=[[1,1],[-1,1]]^{\mathrm {T}}$ and $\Sigma_1=\mathbf{I}$, where $\mu_0$, $\Sigma_0$ and $\mu_1$, $\Sigma_1$ denote the means and covariance matrices of the Gaussian distributions associated with instances in classes $0$ and $1$, respectively. The features $x_{1:2}$ are called $core\_1$ and $core\_2$. Random values are assigned to the features $x_{3:4}$ ($noise\_1$ and $noise\_2$). Such a configuration results in a dataset holding two predictive (important) and two random noise (non-important) features with a small mixture area between the important features to introduce some degree of complexity.
        
        The second synthetic dataset is more robust. It is a 20-dimensional (20-FT) dataset created using the OpenXAI synthetic data generation tool, whose algorithm and properties are described in \citet{agarwal2022openxai}. According to the authors, the algorithm ensures the creation of a dataset that encapsulates feature dependencies and clear local neighborhoods, key properties to guarantee the explanations derived from this synthetic dataset remain consistent with the behavior of the models trained on such data. Using controlled synthetic data for XAI checking meets the correctness dimension of explainability evaluation~\cite{nauta2023from}.
    
        We trained the XRFC and 3H-NN models for the synthetic datasets experiments. The XRFC classifier achieved $86.5\%$ accuracy in the 4-FT data and $83.5\%$ accuracy in the 20-FT data, while the 3H-NN model achieved $97.5\%$ accuracy in the 4-FT data and $83.5\%$ accuracy in the 20-FT synthetic dataset. Evaluating explanation methods on accurate models is good practice in controlled settings based on synthetic data~\cite{nauta2023from}. The higher the model's accuracy level, the more it can be assumed that the model adhered to the reasoning designed in the data.
        
        Note that we are not basing our benchmark experiments on synthetic datasets with massive amounts of instances. We understand that a reasonable amount of data is essential for effectively training and testing machine learning models. According to \citet{aas2021explaining}, the well-known feature importance methods become unstable in tasks with more than ten dimensions. In this sense, data dimensionality is more critical to assessing the stability of feature importance methods than the number of instances. Thus, we generate synthetic datasets with enough instances ($1,000$ for each dataset) to train our models and evaluate XAI methods.
        
        Table~\ref{tab:xrfc_4ft} shows the stability metrics RIS, ROS, and RES for T-Explainer, TreeSHAP, and LIME, explaining the XRFC model on the 4-FT data. Notice that TreeSHAP has the best RIS, ROS, and RES results.
        The good performance of TreeSHAP is expected, as 4-FT is a low-dimensional dataset, and TreeSHAP takes advantage of it by relying on a deterministic version of Shapley values computation~\cite{amparore2021}. 
        Table~\ref{tab:3hnn_4ft} shows the same metrics for the 3H-NN model in the 4-FT dataset. The T-Explainer performed considerably better than the other XAI methods in RIS ($7\times$ better than DeepLIFT), ROS, and the RES metric.
        
        %\begin{table}[!htb]
        %\begin{minipage}{.48\linewidth}
        \begin{table}[!htb]
            \centering 
            \addtolength{\tabcolsep}{-1pt}
            \small{
            \begin{tabular}{lrrrrc}
                \toprule
                XRFC & \multicolumn{2}{c}{RIS} & \multicolumn{2}{c}{ROS} & RES \\ 
                \cmidrule(lr){1-1} \cmidrule(lr){2-3} \cmidrule(lr){4-5} \cmidrule(lr){6-6}
                XAI      & Max            & Mean           & Max            & Mean           & Max \\ 
                \cmidrule(lr){1-1} \cmidrule(lr){2-6}
                T-Exp    & 13,819         & 111.4          & 6e+06          & 38,904         & \textbf{0} \\
                TreeSHAP & \textbf{5,053} & \textbf{38.0}  & \textbf{5e+05} & \textbf{8,226} & \textbf{0} \\
                LIME     & 10,895         & 200.4          & 6e+06          & 86,050         & 3e-04 \\
                \bottomrule
            \end{tabular}}
            \caption{Stability of XAI methods explaining the XRFC predictions on the 4-FT synthetic dataset. T-Exp denotes T-Explainer and Max refers to maximum values.}
            \label{tab:xrfc_4ft}
        \end{table}
        %\end{minipage}\hfill
        %\begin{minipage}{.48\linewidth}
        \begin{table}[!htb]
            \centering 
            \addtolength{\tabcolsep}{-1pt}
            \small{
            \begin{tabular}{lrrrrc}
                \toprule
                3H-NN & \multicolumn{2}{c}{RIS} & \multicolumn{2}{c}{ROS} & RES \\ 
                \cmidrule(lr){1-1} \cmidrule(lr){2-3} \cmidrule(lr){4-5} \cmidrule(lr){6-6}
                XAI      & Max            & Mean           & Max            & Mean           & Max \\ 
                \cmidrule(lr){1-1} \cmidrule(lr){2-6}
                T-Exp         & \textbf{176} & \textbf{5.97} & \textbf{806}  & \textbf{5.82}  & \textbf{0} \\
                SHAP          & 2,010        & 38.59         & 12,122        & 51.39          & \textbf{0} \\
                LIME          & 4,625        & 127.1         & 1.7e+05       & 296.9          & 3e-02 \\
                I-Grad        & 2,465        & 19.33         & 1,169         & 7.41           & \textbf{0} \\
                I$\times$Grad & 1,316        & 9.75          & 1,535         & 7.74           & 1e-05 \\
                DeepLIFT      & 1,316        & 9.75          & 1,535         & 7.74           & 1e-05 \\
                \bottomrule
            \end{tabular}}
            \caption{Stability of XAI methods explaining the 3H-NN predictions on the 4-FT dataset. I-Grad and I$\times$Grad denote Integrated Gradients and Input $\times$ Gradient, respectively.}    
            \label{tab:3hnn_4ft}
        \end{table}
        %\end{minipage}
        %\end{table}

        Table~\ref{tab:faith_3hnn_4ft} presents the PGI results of the explainers applied to the 3H-NN predictions on the 4-FT data. T-Explainer explanations are considerably more faithful than the other methods in identifying the most important feature for predictions (Top 1 column), i.e., T-Explainer's explanations demonstrated the closest behavior to what the underlying model learned as the most important feature. When the task considered more features, the second and third most important, SHAP outperformed the T-Explainer; however, our method performed consistently more faithfully to the model's predictions than the other explainers (columns Top 2 and Top 3).

        \begin{table}[!htb]
            \centering 
            \addtolength{\tabcolsep}{1pt}
            \small{
            \begin{tabular}{lrrr}
                \toprule
                3H-NN & \multicolumn{3}{c}{PGI} \\ 
                \cmidrule(lr){1-1} \cmidrule(lr){2-4}
                XAI           & Top 1           & Top 2           & Top 3  \\ 
                \cmidrule(lr){1-1} \cmidrule(lr){2-4}
                T-Exp         & \textbf{0.7337} & 0.7335          & 0.7334          \\
                SHAP          & 0.6063          & \textbf{0.8402} & \textbf{0.7569} \\
                LIME          & 0.4973          & 0.4960          & 0.4956          \\
                I-Grad        & 0.4973          & 0.4960          & 0.4956          \\
                I$\times$Grad & 0.4973          & 0.4960          & 0.4956          \\
                DeepLIFT      & 0.4973          & 0.4960          & 0.4956          \\
                \bottomrule
            \end{tabular}}
            \caption{Faithfulness of XAI methods explaining the 3H-NN predictions on the 4-FT dataset according to the Top \textit{k} most important features.}
            \label{tab:faith_3hnn_4ft}
        \end{table}
    
        Tables~\ref{tab:xrfc_20ft} and~\ref{tab:3hnn_20ft} represent the stability of the XRFC and 3H-NN models in the 20-FT dataset.
        In Table~\ref{tab:xrfc_20ft}, we observe that the T-Explainer is superior to TreeSHAP and LIME in the RIS and ROS metrics for both the maximum and mean values. 
        TreeSHAP reached a close performance regarding maximum ROS, but T-Explainer is considerably better in average ROS. 
        Similar results can be observed in Table~\ref{tab:3hnn_20ft}, where the T-Explainer outperforms all other explainers in terms of RIS ($3\times$ better than Integrated Gradients), with slightly better performance than Integrated Gradients in the mean ROS but outperforms the other explainers in maximum ROS. 
        
        SHAP was the most unstable method, which can be justified in the context of the 20-FT dataset because, in high-dimensional spaces, SHAP uses a random sampling algorithm rather than a deterministic one, leading to unstable explanations~\cite{amparore2021}. Moreover, SHAP is also prone to extrapolations~\cite{hooker2019please,hooker2021unrestricted}.
        
        %\begin{table}[!htb]
        %\begin{minipage}{.48\linewidth}
        \begin{table}[!htb]
            \centering 
            \addtolength{\tabcolsep}{-1pt}
            \small{
            \begin{tabular}{lrrrrc}
                \toprule
                XRFC & \multicolumn{2}{c}{RIS} & \multicolumn{2}{c}{ROS} & RES \\ 
                \cmidrule(lr){1-1} \cmidrule(lr){2-3} \cmidrule(lr){4-5} \cmidrule(lr){6-6}
                XAI      & Max            & Mean           & Max            & Mean           & Max \\ 
                \cmidrule(lr){1-1} \cmidrule(lr){2-6}
                T-Exp    & \textbf{728}   & \textbf{28.95} & \textbf{2.6e+06} & \textbf{35,596} & \textbf{0} \\
                TreeSHAP & 897            & 37.83          & 3.3e+06          & 64,204          & \textbf{0} \\
                LIME     & 1,117          & 143.6          & 21.6e+06         & 3.5e+05         & 1e-04 \\
                \bottomrule
            \end{tabular}}
            \caption{Stability of XAI methods explaining the XRFC predictions on the 20-FT synthetic dataset.}
            \label{tab:xrfc_20ft}
        \end{table}
        %\end{minipage}\hfill
        %\begin{minipage}{.48\linewidth}
        \begin{table}[!htb]
            \centering 
            \addtolength{\tabcolsep}{-1pt}
            \small{
            \begin{tabular}{lrrrrc}
                \toprule
                3H-NN & \multicolumn{2}{c}{RIS} & \multicolumn{2}{c}{ROS} & RES \\ 
                \cmidrule(lr){1-1} \cmidrule(lr){2-3} \cmidrule(lr){4-5} \cmidrule(lr){6-6}
                XAI      & Max            & Mean           & Max            & Mean           & Max \\ 
                \cmidrule(lr){1-1} \cmidrule(lr){2-6}
                T-Exp         & \textbf{717} & \textbf{26.2} & \textbf{17,887} & \textbf{99.98} & \textbf{0} \\
                SHAP          & 2.8e+05      & 7,272         & 2e+06           & 14,076         & 2e-01 \\
                LIME          & 10,182       & 125.6         & 28,063          & 301.9          & 3e-02 \\
                I-Grad        & 2,279        & 28.35         & 39,287          & 107.4          & \textbf{0} \\
                I$\times$Grad & 11,815       & 63.36         & 60,787          & 217.2          & 5e-06 \\
                DeepLIFT      & 11,813       & 63.36         & 60,781          & 217.2          & 5e-06 \\
                \bottomrule
            \end{tabular}}
            \caption{Stability of XAI methods explaining the 3H-NN predictions on the 20-FT dataset.}
            \label{tab:3hnn_20ft}
        \end{table}
        %\end{minipage}
        %\end{table}
        
        In general, Tables~\ref{tab:xrfc_4ft}, \ref{tab:3hnn_4ft}, \ref{tab:xrfc_20ft}, and \ref{tab:3hnn_20ft} show that the T-Explainer is robust in different stability metrics that evaluate the consistency and continuity of explanations in controlled synthetic data tasks~\cite{nauta2023from}, being among the best performance methods, especially when the data dimensionality is high.

        Observing the PGI results in Table~\ref{tab:faith_3hnn_20ft}, we have the T-Explainer as the most faithful explanation method for the predictive behavior of the original model for the most important feature (Top 1 column). Enlarging the set of the most important $k$ features (Top 3--9 columns), SHAP presented the best performances; however, T-Explainer consistently outperformed all the other explainers, being the second most faithful explanation method explaining the 3H-NN predictions on the 20-FT synthetic dataset.

        \begin{table}[!htb]
            \centering 
            \addtolength{\tabcolsep}{0pt}
            \small{
            \begin{tabular}{lrrrrr}
                \toprule
                3H-NN & \multicolumn{5}{c}{PGI} \\ 
                \cmidrule(lr){1-1} \cmidrule(lr){2-6}
                XAI           & Top 1           & Top 3           & Top 5           & Top 7           & Top 9  \\ 
                \cmidrule(lr){1-1} \cmidrule(lr){2-6}
                T-Exp         & \textbf{0.5913} & 0.5738          & 0.5674          & 0.5624          & 0.5669 \\
                SHAP          & 0.5844          & \textbf{0.6540} & \textbf{0.6467} & \textbf{0.6580} & \textbf{0.6724} \\
                LIME          & 0.4586          & 0.4548          & 0.4536          & 0.4536          & 0.4535 \\
                I-Grad        & 0.4562          & 0.4540          & 0.4538          & 0.4536          & 0.4536 \\
                I$\times$Grad & 0.5604          & 0.5621          & 0.5501          & 0.5482          & 0.5504 \\
                DeepLIFT      & 0.5604          & 0.5621          & 0.5501          & 0.5482          & 0.5504 \\
                \bottomrule
            \end{tabular}}
            \caption{Faithfulness of XAI methods explaining the 3H-NN predictions on the 20-FT dataset according to the Top \textit{k} most important features.}
            \label{tab:faith_3hnn_20ft}
        \end{table}

\begin{figure*}[!htb]
            \centering
            \begin{subfigure}[b]{\textwidth}
                \centering
                \includegraphics[width=0.9\linewidth]{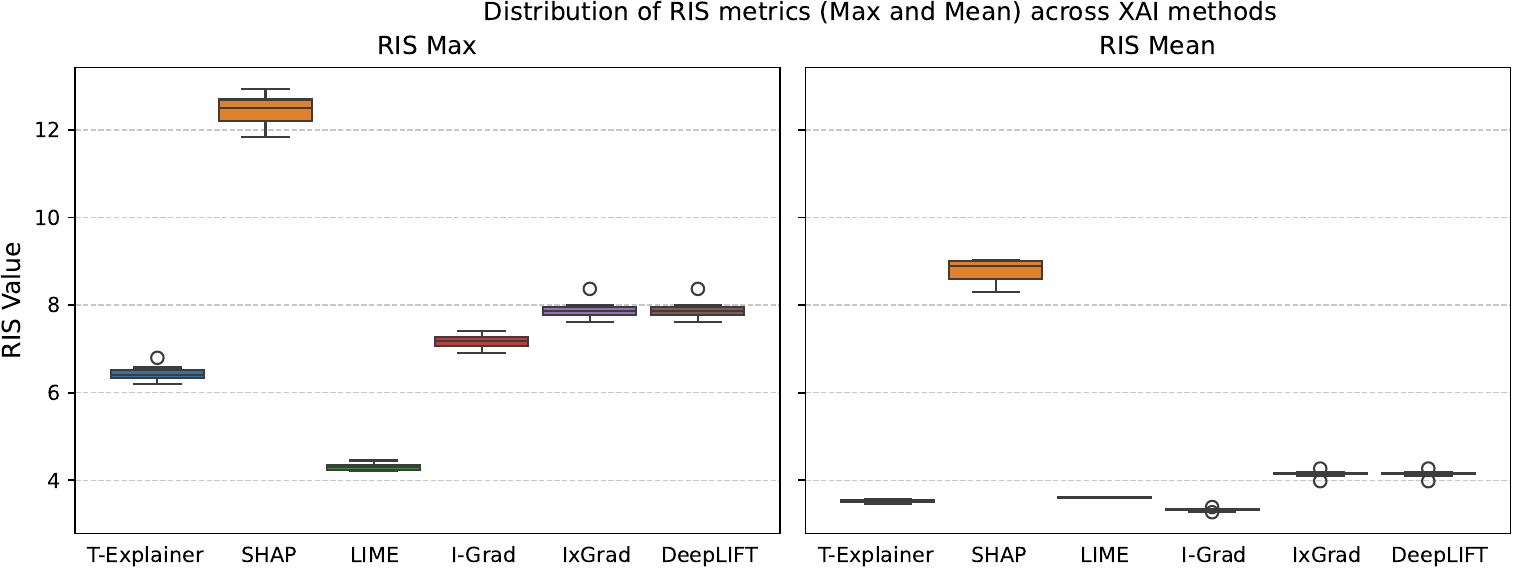}
                \caption{RIS.}
                \label{fig:ris_dist}
                \vspace{1em}
            \end{subfigure}
            \begin{subfigure}[b]{\textwidth}
                \centering
                \includegraphics[width=0.9\linewidth]{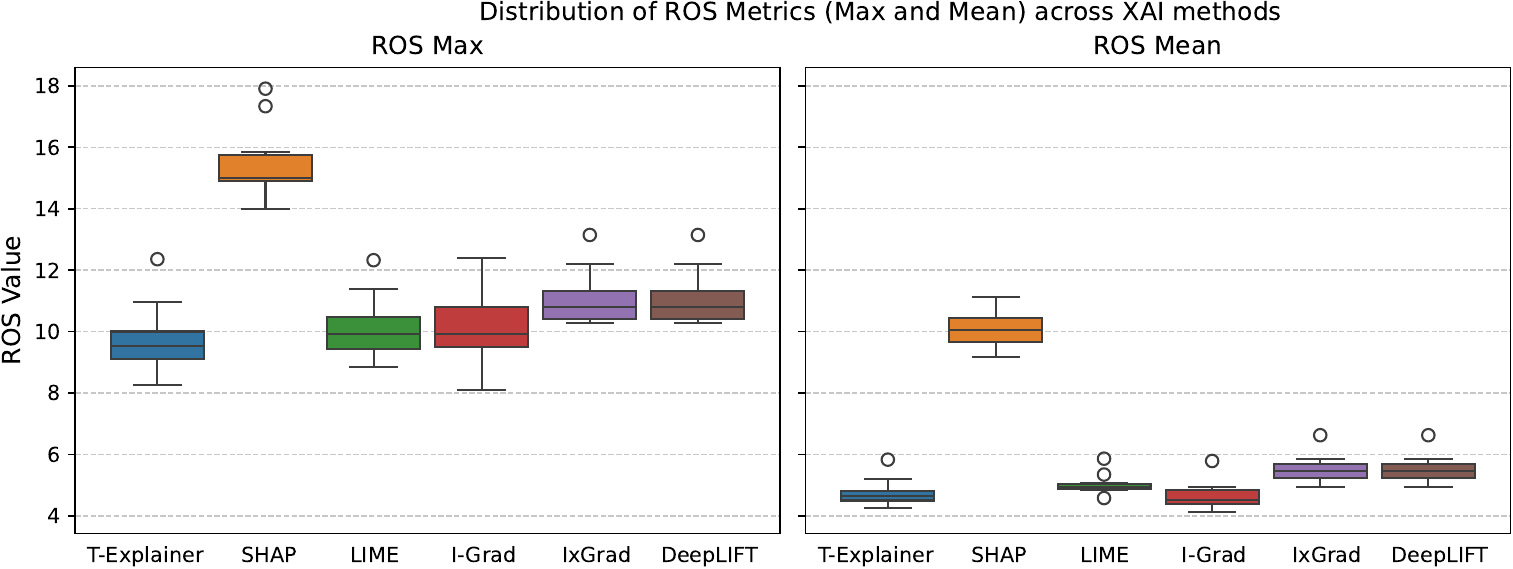}
                \caption{ROS.}
                \label{fig:ros_dis}
            \end{subfigure}
            \caption{Behavior of XAI methods explaining samples predicted with the 3H-NN model trained on the 20-FT dataset when running RIS and ROS metrics multiple times.
            The \textit{y}-axes are on a logarithmic scale for better visualization. }
            \label{fig:dists}
        \end{figure*}

        Figure~\ref{fig:dists} compares the RIS and ROS metrics. Once they are based on normal perturbations, we executed both metrics multiple times to evaluate the behavior of the XAI explainers and add statistical significance to our empirical experiments. We randomly selected $100$ samples from the 20-FT data (fixing the same seed throughout the experiment for reproducibility) and performed RIS and ROS metrics $10$ times each, evaluating explanations of the 3H-NN model. Similarly to previous experiments, we present the maximum and mean values of RIS and ROS over the runs.
        The boxplots in Figure~\ref{fig:dists} show that T-Explainer is ranked among the most stable approaches, following the earlier findings presented in Table~\ref{tab:3hnn_20ft}.

        In summary, T-Explainer's performance when dealing with the XRFC classifier, a non-differentiable model, was remarkable for more complex tasks (e.g., explaining predictions for the 20-FT data). Such performances have significant practical implications, suggesting that the T-Explainer might perform well even when applied to generate explanations for non-differentiable machine learning models. The following section presents the evaluation results on real-world data.

%-----------------------------------------------------------------------------
    \subsection{Real-world Data}
    \label{experiments:real}
        This Section extends the evaluations and comparisons by applying real-world data from different domains. Specifically, we run experiments using four well-known real-world datasets with distinct properties regarding dimensionality, the presence of categorical attributes, and size. 
        
        The Banknote Authentication~\cite{banknote_authentication} is a $4$-dimensional dataset containing $1,372$ instances with measurements of genuine and forged banknote specimens. 
        The German Credit~\cite{german_credit_data} comprises $9$ features (numerical and categorical) covering financial, demographic, and personal information from $1,000$ credit applicants, each categorized into good or bad risk. 
        
        The Home Equity Line of Credit (HELOC) dataset provided by FICO~\cite{heloc} consists of financial attributes from anonymized applications for home equity lines of credit submitted by $9,871$ real homeowners. The task in the HELOC dataset is to predict whether an applicant has a good or bad risk of repaying the HELOC account within two years. 
        The largest dataset in our study is HIGGS~\cite{baldi2014searching}, which contains $28$ features about simulated collision events to distinguish between Higgs bosons signals and a background process. The original HIGGS dataset contains $11$ million instances, but we used the $98,050$ instances version available at OpenML~\cite{vanschoren2014openml}. 
        Table~\ref{tab:real_ds} summarizes the real datasets. 
    
        \begin{table}[!htb]
            \centering 
            \addtolength{\tabcolsep}{-1pt}
            \small{
            \begin{tabular}{lrrrrl}
                \toprule
                                & Banknote & German & HELOC & HIGGS  \\ \midrule
                \#Instances     & 1,372    & 1,000  & 9,871 & 98,050 \\
                \#Num features  & 4        & 4      & 21    & 28     \\
                \#Cat features  & 0        & 5      & 2     & 0      \\
                \#Classes       & 2        & 2      & 2     & 2      \\
                \bottomrule
            \end{tabular}}
            \caption{Properties of the real datasets. Num features and Cat features refer to numeric and categorical features, respectively.}
            \label{tab:real_ds}
        \end{table}
    
        The experiments with real-world data were conducted using only the neural network models, as some of the explanation models with which T-Explainer is compared are specifically designed for neural networks. Moreover, the neural network models are differentiable, thus meeting the theoretical requirements that support the T-Explainer technique.
        We further discuss tree-based models in Section~\ref{discussions}.
     
        For experiments with Banknote Authentication and German Credit datasets, we selected the 3H-NN model.
        In Table~\ref{tab:3hnn_bba}, one can see that T-Explainer is less stable than Integrated Gradients, but it is the second-best method on average for RIS and ROS (tied with Integrated Gradients for RES).  

        \begin{table}[!htb]
            \centering 
            \addtolength{\tabcolsep}{-1pt}
            \small{
            \begin{tabular}{lrrrrc}
                \toprule
                3H-NN & \multicolumn{2}{c}{RIS} & \multicolumn{2}{c}{ROS} & RES \\ 
                \cmidrule(lr){1-1} \cmidrule(lr){2-3} \cmidrule(lr){4-5} \cmidrule(lr){6-6}
                XAI      & Max            & Mean           & Max            & Mean           & Max \\ 
                \cmidrule(lr){1-1} \cmidrule(lr){2-6}
                T-Exp         & 175.29         & 3.09          & 429.31          & 4.05          & \textbf{0} \\
                SHAP          & 1.3e+05        & 468           & 1.9e+05         & 543           & \textbf{0} \\
                LIME          & 97.57          & 9.34          & 849.02          & 13.43         & 3e-02 \\
                I-Grad        & \textbf{12.37} & \textbf{1.54} & 429.23          & \textbf{2.09} & \textbf{0} \\
                I$\times$Grad & 175.33         & 4.41          & \textbf{427.97} & 4.12          & 2e-05 \\
                DeepLIFT      & 175.33         & 4.41          & \textbf{427.97} & 4.12          & 2e-05 \\
                \bottomrule
            \end{tabular}}
            \caption{Stability of XAI methods explaining the 3H-NN predictions on the Banknote Authentication dataset.}
            \label{tab:3hnn_bba}
        \end{table}

        Analyzing the PGI results in Table~\ref{tab:faith_3hnn_bba}, we see that all the methods are tied to a faithful behavior between explanations and model predictions according to the most important feature identified in their explanations. However, T-Explainer's performance grows when considering the second and third most important features. Only SHAP and T-Explainer achieved a faithfulness higher than $0.5$ for their explanations' three most important features. In a low-dimensional (four features, see Table~\ref{tab:real_ds}) set such as the Banknote Authentication, it is possible to conjecture that an explainer with a PGI below $0.5$ generally lacks a highly significant feature in its explanations.

        \begin{table}[!htb]
            \centering 
            \addtolength{\tabcolsep}{1pt}
            \small{
            \begin{tabular}{lrrr}
                \toprule
                3H-NN & \multicolumn{3}{c}{PGI} \\ 
                \cmidrule(lr){1-1} \cmidrule(lr){2-4}
                XAI           & Top 1           & Top 2           & Top 3  \\ 
                \cmidrule(lr){1-1} \cmidrule(lr){2-4}
                T-Exp         & 0.4514          & \textbf{0.7304} & \textbf{0.9563} \\
                SHAP          & 0.4514          & 0.4514          & 0.4464          \\
                LIME          & 0.4514          & 0.4514          & 0.6634          \\
                I-Grad        & 0.4514          & 0.4514          & 0.4514          \\
                I$\times$Grad & 0.4514          & 0.4514          & 0.4514          \\
                DeepLIFT      & 0.4514          & 0.4514          & 0.4514          \\
                \bottomrule
            \end{tabular}}
            \caption{Faithfulness of XAI methods explaining the 3H-NN predictions on the Banknote Authentication dataset according to the Top \textit{k} most important features.}
            \label{tab:faith_3hnn_bba}
        \end{table}
        
        Table~\ref{tab:3hnn_ger} presents the stability tests for the predictions of the 3H-NN model on the German Credit data. 
        The original version of the dataset has $20$ numerical and categorical attributes. However, applying it directly to a machine learning task is difficult due to its intricate categorization system. 
        We preprocessed the German Credit data to clean it, resulting in a reduced version holding nine features ($4$ numerical and $5$ categorical features, see Table~\ref{tab:real_ds}). 
        As shown in Table~\ref{tab:3hnn_ger}, the T-Explainer performed as the most stable explainer regarding RIS and maximum ROS perturbations, being the second best for the mean ROS. 
        
        \begin{table}[!htb]
            \centering 
            \addtolength{\tabcolsep}{-1pt}
            \small{
            \begin{tabular}{lrrrrc}
                \toprule
                3H-NN & \multicolumn{2}{c}{RIS} & \multicolumn{2}{c}{ROS} & RES \\ 
                \cmidrule(lr){1-1} \cmidrule(lr){2-3} \cmidrule(lr){4-5} \cmidrule(lr){6-6}
                XAI      & Max            & Mean           & Max            & Mean           & Max \\ 
                \cmidrule(lr){1-1} \cmidrule(lr){2-6}
                T-Exp         & \textbf{1,971} & \textbf{49.4} & \textbf{15,735} & 239.9          & \textbf{0}\\
                SHAP          & 1.0e+06        & 14,044        & 2.5e+07         & 62,883         & 2.1e-01 \\
                LIME          & 7,547          & 131.1         & 1.8e+05         & 739.9          & 2.4e-02 \\
                I-Grad        & 8,709          & 59.5          & 58,721          & \textbf{185.7} & \textbf{0} \\
                I$\times$Grad & 10,933         & 63.2          & 1.0e+05         & 350.8          & 2.8e-06 \\
                DeepLIFT      & 10,934         & 63.2          & 1.0e+05         & 350.8          & 2.4e-06 \\
                \bottomrule
            \end{tabular}}
            \caption{Stability of XAI methods explaining the 3H-NN predictions on the German Credit dataset.}
            \label{tab:3hnn_ger}
        \end{table}

        Table~\ref{tab:faith_3hnn_ger} shows close performances regarding the PGI metric among all feature importance methods for 3H-NN predictions on German Credit data. LIME, the T-Explainer, and SHAP were the most faithful explainers in the incremental verification process of the top $k$ features.

        \begin{table}[!htb]
            \centering 
            \addtolength{\tabcolsep}{0pt}
            \small{
            \begin{tabular}{lrrrrr}
                \toprule
                3H-NN & \multicolumn{5}{c}{PGI} \\ 
                \cmidrule(lr){1-1} \cmidrule(lr){2-6}
                XAI           & Top 1           & Top 3           & Top 5           & Top 7           & Top 9  \\ 
                \cmidrule(lr){1-1} \cmidrule(lr){2-6}
                T-Exp         & 0.8044          & 0.8164          & \textbf{0.8311} & 0.8198          & 0.8178 \\
                SHAP          & 0.8245          & 0.8193          & 0.8215          & \textbf{0.8242} & \textbf{0.8240} \\
                LIME          & \textbf{0.8396} & \textbf{0.8231} & 0.8196          & 0.8184          & 0.8187 \\
                I-Grad        & 0.8257          & 0.8182          & 0.8156          & 0.8159          & 0.8159 \\
                I$\times$Grad & 0.8149          & 0.8212          & 0.8222          & 0.8220          & 0.8220 \\
                DeepLIFT      & 0.8149          & 0.8212          & 0.8222          & 0.8220          & 0.8220 \\
                \bottomrule
            \end{tabular}}
            \caption{Faithfulness of XAI methods explaining the 3H-NN predictions on the German Credit dataset according to the Top \textit{k} most important features.}
            \label{tab:faith_3hnn_ger}
        \end{table}
    
        Tables~\ref{tab:3hnn_bba} and~\ref{tab:3hnn_ger} indicate that gradient-based methods are more stable to input/output perturbations than SHAP and, in some cases, also than LIME. As one can notice, the T-Explainer competes quite well with other gradient-based methods regarding stable explanations. Although Tables~\ref{tab:faith_3hnn_bba} and \ref{tab:faith_3hnn_ger} place the T-Explainer between the better feature importance methods for faithful explanations.
    
        HELOC and HIGGS are more robust datasets in terms of dimensionality and size. We selected the 3H-NN and the five-layer neural network models to compare T-Explainer with the other explainers on the real-world datasets. Specifically, we trained the 3H-NN and 5H-128-NN classifiers in HELOC, achieving $71.34\%$ and $73.27\%$ of accuracy, respectively. Despite the accuracy values being lower than those we achieved before, these performances align with the results reported in the literature~\cite{borisov2022deep}. 
        
        Table~\ref{tab:3hnn_hel} shows that Integrated Gradients is the most stable explainer for almost all metrics (behind LIME only for mean RIS but with very close numbers). However, we highlight the performance of T-Explainer, especially regarding the metrics' means. Only Integrated Gradients, LIME, and T-Explainer kept the mean RIS under ten units, which can be taken as virtually the same mean stability to input perturbations. Similar results are observed in mean ROS, where Integrated Gradients, LIME, and T-Explainer achieved the smallest values, with T-Explainer the second best. 
        Table~\ref{tab:5hnn_hel} shows the stability of the explainers when applied to the five-layer neural network. T-Explainer was the most stable method, with Integrated Gradients being the second best for mean RIS and LIME as the second best for mean ROS. Only the T-Explainer and Integrated Gradients achieved the highest levels of stability in RES.
    
        %\begin{table}[!htb]
        %\begin{minipage}{.49\linewidth}
        \begin{table}[!htb]
            \centering 
            \addtolength{\tabcolsep}{-1pt}
            \small{
            \begin{tabular}{lrrrrc}
                \toprule
                3H-NN & \multicolumn{2}{c}{RIS} & \multicolumn{2}{c}{ROS} & RES \\ 
                \cmidrule(lr){1-1} \cmidrule(lr){2-3} \cmidrule(lr){4-5} \cmidrule(lr){6-6}
                XAI      & Max            & Mean           & Max            & Mean           & Max \\ 
                \cmidrule(lr){1-1} \cmidrule(lr){2-6}
                T-Exp         & 1,443          & 8.49          & 94,943          & 527.9          & \textbf{0} \\
                SHAP          & 84,794         & 1,330         & 1.7e+07         & 64,693         & 1.5e-02 \\
                LIME          & 509.7          & \textbf{3.84} & 3.7e+05         & 626.5          & 9.3e-02 \\
                I-Grad        & \textbf{159.8} & 4.21          & \textbf{85,463} & \textbf{401.1} & \textbf{0} \\
                I$\times$Grad & 3,754          & 33.67         & 3.5e+05         & 1,879          & 8.7e-07 \\
                DeepLIFT      & 3,749          & 33.64         & 3.5e+05         & 1,878          & 9.0e-07 \\
                \bottomrule
            \end{tabular}}
            \caption{Stability of XAI methods explaining the 3H-NN predictions on the HELOC dataset.}
            \label{tab:3hnn_hel}
        \end{table}
        %\end{minipage}\hfill
        %\begin{minipage}{.49\linewidth}
        \begin{table}[!htb]
            \centering 
            \addtolength{\tabcolsep}{-1pt}
            \small{
            \begin{tabular}{lrrrrc}
                \toprule
                5H-128-NN & \multicolumn{2}{c}{RIS} & \multicolumn{2}{c}{ROS} & RES \\ 
                \cmidrule(lr){1-1} \cmidrule(lr){2-3} \cmidrule(lr){4-5} \cmidrule(lr){6-6}
                XAI      & Max            & Mean           & Max            & Mean           & Max \\ 
                \cmidrule(lr){1-1} \cmidrule(lr){2-6}
                T-Exp         & \textbf{154.0} & \textbf{5.37} & \textbf{1.4e+05} & \textbf{362.5} & \textbf{0} \\
                SHAP          & 52,924         & 1,757         & 1.6e+07          & 67,451         & 1.9e-01 \\
                LIME          & 5,083          & 39.96         & 2.5e+05          & 935.0          & 1.6e-02 \\
                I-Grad        & 255.7          & 7.21          & 1.9e+07          & 19,283         & \textbf{0} \\
                I$\times$Grad & 11,769         & 66.01         & 4.3e+05          & 3,215          & 1.8e-06 \\
                DeepLIFT      & 11,785         & 66.06         & 4.3e+05          & 3,218          & 1.7e-06 \\
                \bottomrule
            \end{tabular}}
            \caption{Stability of XAI methods explaining the 5H-128-NN predictions on the HELOC dataset.}
            \label{tab:5hnn_hel}
        \end{table}
        %\end{minipage}
        %\end{table}

        Tables~\ref{tab:faith_3hnn_hel} and \ref{tab:faith_5hnn_hel} present the PGI performance of the explainers. All methods increased their faithfulness values by including more features in the set of top $k$ features, indicating that the HELOC dataset has multiple significant features for the underlying model. For the 3H-NN model, the T-Explainer presented the highest PGI value considering the nine most essential features, i.e., our method could identify nine important features with high fidelity to the predictions of the model out of more than $20$ predictive features. For the 5H-128-NN model, T-Explainer performed better up to the three most essential features, although our method was the second with higher PGI, after Integrated Gradients, the most faithful explainer to larger top $k$ feature sets.

        \begin{table}[!htb]
            \centering 
            \addtolength{\tabcolsep}{0pt}
            \small{
            \begin{tabular}{lrrrrr}
                \toprule
                3H-NN & \multicolumn{5}{c}{PGI} \\ 
                \cmidrule(lr){1-1} \cmidrule(lr){2-6}
                XAI           & Top 1           & Top 3           & Top 5           & Top 7           & Top 9  \\ 
                \cmidrule(lr){1-1} \cmidrule(lr){2-6}
                T-Exp         & 0.6360          & 0.7364          & 0.8146          & 0.8595          & \textbf{0.8712} \\
                SHAP          & 0.6309          & 0.6529          & 0.7268          & 0.7740          & 0.7995 \\
                LIME          & 0.6274          & 0.6543          & 0.8350          & 0.8424          & 0.8449 \\
                I-Grad        & \textbf{0.6379} & \textbf{0.7774} & \textbf{0.8557} & \textbf{0.8614} & 0.8622 \\
                I$\times$Grad & \textbf{0.6379} & 0.7710          & 0.8436          & 0.8539          & 0.8419 \\
                DeepLIFT      & \textbf{0.6379} & 0.7710          & 0.8436          & 0.8539          & 0.8419 \\
                \bottomrule
            \end{tabular}}
            \caption{Faithfulness of XAI methods explaining the 3H-NN predictions on the HELOC dataset according to the Top \textit{k} most important features.}
            \label{tab:faith_3hnn_hel}
        \end{table}

        \begin{table}[!htb]
            \centering 
            \addtolength{\tabcolsep}{0pt}
            \small{
            \begin{tabular}{lrrrrr}
                \toprule
                5H-128-NN & \multicolumn{5}{c}{PGI} \\ 
                \cmidrule(lr){1-1} \cmidrule(lr){2-6}
                XAI           & Top 1           & Top 3           & Top 5           & Top 7           & Top 9  \\ 
                \cmidrule(lr){1-1} \cmidrule(lr){2-6}
                T-Exp         & \textbf{0.7028} & \textbf{0.7454} & 0.7470          & 0.7689          & 0.7821 \\
                SHAP          & 0.6373          & 0.6617          & 0.6948          & 0.7329          & 0.7667 \\
                LIME          & 0.6627          & 0.7044          & \textbf{0.8022} & 0.7223          & 0.7663 \\
                I-Grad        & 0.6763          & 0.7130          & 0.7617          & \textbf{0.7923} & \textbf{0.7910} \\
                I$\times$Grad & 0.6506          & 0.6847          & 0.7198          & 0.7445          & 0.7566 \\
                DeepLIFT      & 0.6506          & 0.6847          & 0.7198          & 0.7445          & 0.7566 \\
                \bottomrule
            \end{tabular}}
            \caption{Faithfulness of XAI methods explaining the 5H-128-NN predictions on the HELOC dataset according to the Top \textit{k} most important features.}
            \label{tab:faith_5hnn_hel}
        \end{table}
    
        We trained the 3H-NN and 5H-64-NN classifiers on HIGGS data, which achieved accuracy $69.58\%$ and $66.83\%$, respectively, close to the values reported in previous work~\cite{borisov2022deep}. We included a set of benchmarks using a five-layer neural network here to maintain some similarity to the architectures proposed in \citet{baldi2014searching}, which explored the use of deep networks in HIGGS. 
        Tables~\ref{tab:3hnn_hig} and~\ref{tab:5hnn_hig} show good stability performances of gradient-based feature importance explainers. T-Explainer outperforms them in RIS perturbations and remains competitive in ROS and RES metrics. Such results demonstrate the ability of the T-Explainer to generate consistent explanations through different levels of model complexity and data dimensionality. 
        Note that SHAP is more unstable than LIME, with SHAP's claimed advantageous performance predominantly seen in low-dimensional datasets~\cite{amparore2021}.
    
        %\begin{table}[!htb]
        %\begin{minipage}{.49\linewidth}
        \begin{table}[!htb]
            \centering 
            \addtolength{\tabcolsep}{-1pt}
            \small{
            \begin{tabular}{lrrrrc}
                \toprule
                3H-NN & \multicolumn{2}{c}{RIS} & \multicolumn{2}{c}{ROS} & RES \\ 
                \cmidrule(lr){1-1} \cmidrule(lr){2-3} \cmidrule(lr){4-5} \cmidrule(lr){6-6}
                XAI      & Max            & Mean           & Max            & Mean           & Max \\ 
                \cmidrule(lr){1-1} \cmidrule(lr){2-6}
                T-Exp         & \textbf{1,063} & \textbf{45.0} & 69,645          & 304.4          & \textbf{0} \\
                SHAP          & 1.4e+05        & 5,213         & 8.8e+05         & 22,190         & 2.05e-01 \\
                LIME          & 4,424          & 90.0          & 69,458          & 492.6          & 3.19e-02 \\
                I-Grad        & 3,833          & 49.7          & \textbf{27,719} & \textbf{272.4} & \textbf{0} \\
                I$\times$Grad & 2,030          & 79.9          & 70,159          & 604.3          & 6.62e-06 \\
                DeepLIFT      & 2,031          & 79.9          & 70,153          & 604.3          & 6.32e-06 \\
                \bottomrule
            \end{tabular}}
            \caption{Stability of XAI methods explaining the 3H-NN predictions on the HIGGS dataset.}
            \label{tab:3hnn_hig}
        \end{table}
        %\end{minipage}\hfill
        %\begin{minipage}{.49\linewidth}
        \begin{table}[!htb]
            \centering 
            \addtolength{\tabcolsep}{-1pt}
            \small{
            \begin{tabular}{lrrrrc}
                \toprule
                5H-64-NN & \multicolumn{2}{c}{RIS} & \multicolumn{2}{c}{ROS} & RES \\ 
                \cmidrule(lr){1-1} \cmidrule(lr){2-3} \cmidrule(lr){4-5} \cmidrule(lr){6-6}
                XAI      & Max            & Mean           & Max            & Mean           & Max \\ 
                \cmidrule(lr){1-1} \cmidrule(lr){2-6}
                T-Exp         & \textbf{1,359} & \textbf{57.7} & 65,999          & \textbf{330.4} & \textbf{0} \\
                SHAP          & 1.3e+05        & 4,309         & 7.3e+06         & 19,999         & 2.22e-01 \\
                LIME          & 9,582          & 130.8         & 46,700          & 477.2          & 3.50e-02 \\
                I-Grad        & 15,619         & 101.1         & \textbf{37,091} & 452.2          & \textbf{0} \\
                I$\times$Grad & 4,036          & 123.9         & 84,022          & 558.6          & 1.67e-05 \\
                DeepLIFT      & 4,036          & 123.9         & 84,021          & 558.6          & 1.69e-05 \\
                \bottomrule
            \end{tabular}}
            \caption{Stability of XAI methods explaining the 5H-64-NN predictions on the HIGGS dataset.}
            \label{tab:5hnn_hig}
        \end{table}
        %\end{minipage}
        %\end{table}

        Tables~\ref{tab:faith_3hnn_hig} and \ref{tab:faith_5hnn_hig} present the PGI performance of the explainers in the HIGGS dataset. LIME achieved the highest PGI values when explaining predictions made by the 3H-NN and 5H-64-NN neural networks. However, T-Explainer is the second best in describing faithful explanations when considering larger top $k$ feature sets (see Top 7 and 9 columns).

        \begin{table}[!htb]
            \centering 
            \addtolength{\tabcolsep}{0pt}
            \small{
            \begin{tabular}{lrrrrr}
                \toprule
                3H-NN & \multicolumn{5}{c}{PGI} \\ 
                \cmidrule(lr){1-1} \cmidrule(lr){2-6}
                XAI           & Top 1           & Top 3           & Top 5           & Top 7           & Top 9  \\ 
                \cmidrule(lr){1-1} \cmidrule(lr){2-6}
                T-Exp         & 0.6322          & 0.6291          & 0.6395          & 0.6297          & 0.6414 \\
                SHAP          & 0.6355          & 0.6111          & 0.6007          & 0.5920          & 0.5955 \\
                LIME          & \textbf{0.7155} & \textbf{0.7075} & \textbf{0.6841} & \textbf{0.6877} & \textbf{0.6827} \\
                I-Grad        & 0.6380          & 0.5921          & 0.5935          & 0.6071          & 0.6073 \\
                I$\times$Grad & 0.6704          & 0.6367          & 0.6236          & 0.6231          & 0.6351 \\
                DeepLIFT      & 0.6704          & 0.6367          & 0.6236          & 0.6231          & 0.6351 \\
                \bottomrule
            \end{tabular}}
            \caption{Faithfulness of XAI methods explaining the 3H-NN predictions on the HIGGS dataset according to the Top \textit{k} most important features.}
            \label{tab:faith_3hnn_hig}
        \end{table}

        \begin{table}[!htb]
            \centering 
            \addtolength{\tabcolsep}{0pt}
            \small{
            \begin{tabular}{lrrrrr}
                \toprule
                5H-64-NN & \multicolumn{5}{c}{PGI} \\ 
                \cmidrule(lr){1-1} \cmidrule(lr){2-6}
                XAI           & Top 1           & Top 3           & Top 5           & Top 7           & Top 9  \\ 
                \cmidrule(lr){1-1} \cmidrule(lr){2-6}
                T-Exp         & 0.6094          & 0.5923          & 0.6043          & 0.6335          & 0.6357 \\
                SHAP          & 0.5664          & 0.5838          & 0.5941          & 0.6029          & 0.6292 \\
                LIME          & \textbf{0.6768} & \textbf{0.6769} & \textbf{0.6724} & \textbf{0.6545} & \textbf{0.6510} \\
                I-Grad        & 0.6441          & 0.6074          & 0.5897          & 0.5833          & 0.5916 \\
                I$\times$Grad & 0.6221          & 0.6109          & 0.6074          & 0.6228          & 0.6215 \\
                DeepLIFT      & 0.6221          & 0.6109          & 0.6074          & 0.6228          & 0.6215 \\
                \bottomrule
            \end{tabular}}
            \caption{Faithfulness of XAI methods explaining the 5H-64-NN predictions on the HIGGS dataset according to the Top \textit{k} most important features.}
            \label{tab:faith_5hnn_hig}
        \end{table}
        
        Finally, the experiments above show that T-Explainer generally performs well, clearly outperforming well-known model-agnostic methods such as SHAP (and LIME for most tests). 
        %Our method presented competitive results in setups differing from Neural Networks, proving its agnostic capability.
        Moreover, T-Explainer also proved quite competitive with model-specific techniques such as gradient-based ones, thus becoming a new and valuable alternative method for explaining black-box models' predictions.

%=============================================================================
\section{Computational Performance}
\label{performance}
    Another challenge that XAI methods have to face is related to computational performance. Shapley-based approaches like SHAP apply sampling approximations because the exact version of the method is an NP-hard problem with exponential computing time regarding the number of features~\cite{shap_lundberg2017unified}. SHAP is a leading method in feature attribution tasks due to its axiomatic properties, with a range of explainers based on different algorithms, from the exact version to optimized approaches. However, according to its documentation, the most precise algorithm is only feasible for modelings that are nearly limited to $15$ features\footnote{\href{https://shap.readthedocs.io/en/latest/generated/shap.ExactExplainer.html}{https://shap.readthedocs.io/ExactExplainer}, visited in July 2024}.

\begin{figure}[!htb]
        \centering
        \includegraphics[width=1\linewidth]{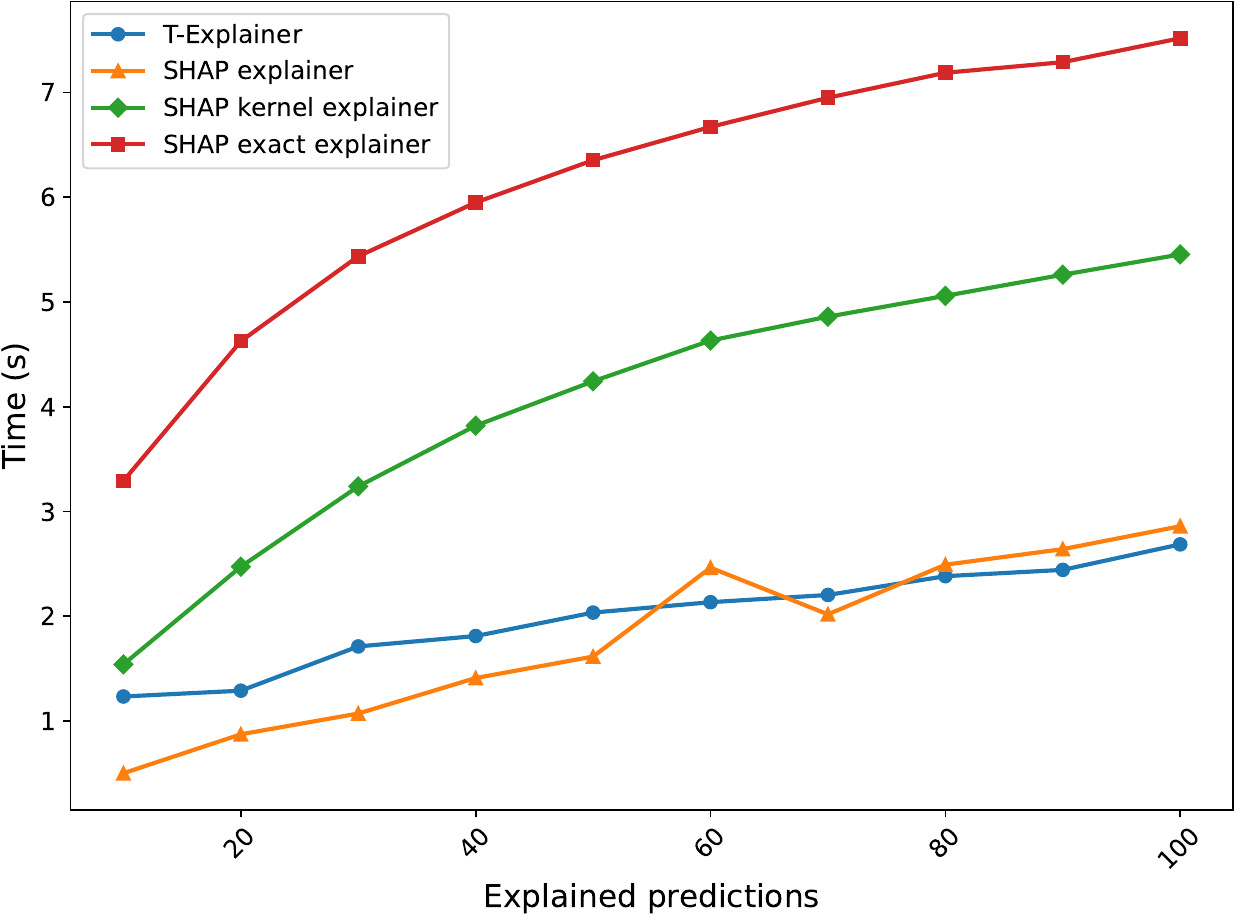}
        \caption{Running times (in log scale) of T-Explainer and SHAP explainers applied to the 3H-NN model trained on 16-dimensional synthetic data.}
        \label{fig:tempos}
    \end{figure}

    For a computational comparison of the performance of the T-Explainer, we generated a synthetic dataset with $16$ features (16-FT) using OpenXAI, similar to the 20-FT dataset we used before (see the Synthetic Data Section). 
    The 3H-NN model was trained with the 16-FT dataset according to the experimental setup defined in Section~\ref{experiments:setup}. 
    We then defined ten explanation tasks, increasing the number of explanations by ten instances from the 16-FT data for each task, ranging from ten to one hundred explanations, to compare the T-Explainer with three implementations of SHAP -- the optimized explainer (the same method used in the stability experiments of Section~\ref{experiments}), KernelSHAP~\cite{shap_lundberg2017unified}, the exact explainer based on standard Shapley values computation. 
    Figure~\ref{fig:tempos} presents the computational performance of T-Explainer and SHAP. The $y$-axis (time) of Figure~\ref{fig:tempos} is on a logarithmic scale for better visualization purposes.
    
    SHAP explainer was the fastest for the first five iterations (explaining sets with 10 to 50 instances). However, the method was the most unstable, as we demonstrated in the stability evaluations. 
    However, the performance of the T-Explainer is quite close to that of the SHAP explainer, and it performs faster than it for sets with 80 instances and above.
    KernelSHAP was slightly more expensive for small sets of instances, while the SHAP exact explainer running time has grown exponentially. 
    In such a performance test, we used a reduced dataset instead of the 20-FT data because applying the SHAP exact explainer in tasks with more than $16$ features is currently impossible. 
    Most machine learning problems involve high-dimensional data. For example, HIGGS is a 28-dimensional dataset with over $98$ thousand instances. With that dataset, it is impossible to apply the exact version of SHAP, and the optimized version of SHAP performs very poorly. 
    Something similar could be said about the HELOC data, which recently motivated a challenge by FICO~\cite{heloc}. 
    In this sense, T-Explainer is placed as a highly competitive XAI approach regarding computational resources.

%=============================================================================
\section{Discussion, Limitations, and Future Work}
\label{discussions}
    T-Explainer is backed by a clearly defined deterministic optimization process for calculating partial derivatives. 
    This new approach gives T-Explainer more robustness, as demonstrated in our results.
    
    Our experiments have focused on binary classification models, which are known to perform well in supervised learning~\cite{borisov2022deep}.
    However, there is no constraint in applying T-Explainer to regression problems or more complex models, such as deep learning models.
    The T-Explainer naturally supports regression or multiclass classification with minor adaptations, which we are currently working on. 
    
    Although T-Explainer has demonstrated competitive results when applied to tree-based models on synthetic data, tree-based architectures impose extra challenges on XAI. 
    Tree classifiers have non-continuous architectures in which constant values are stored in leaf nodes, rendering gradient-based methods inappropriate. 
    The current version of T-Explainer is not yet fully developed to support tree-based models. We are currently designing a mechanism that enables the computation of gradients in tree-based models, extending T-Explainer to operate in a more general context. 
    
    The present implementation of the T-Explainer can handle categorical attributes. It uses an approach that discretizes categorical data into intervals using continuous perturbations, enabling the computation of partial derivatives. 
    However, the approach requires a model retraining to fit the continuous intervals induced in the columns containing one-hot encoded categorical attributes. 
    Retraining introduces additional computational complexity. 
    Getting around the retraining issue is another improvement that we are currently working on. 
    An alternative approach we are considering in this context is to apply the target encoding transformer~\cite{banachewicz2022kaggle}, which converts each nominal value of a categorical attribute into its corresponding expected value.

    Another aspect under improvement is the $\mathbf{h}$ optimization module. To our knowledge, there is no closed solution to determine an optimum value of $\mathbf{h}$ in this context.
    During our investigations, we observed that the finite difference computation presents a certain instability for instances near the model's decision boundaries due to discontinuities imposed by the decision boundaries. 
    The literature on numerical methods offers many alternatives to deal with discontinuities when approximating derivatives through finite differences~\cite{towers2009finite,scheinberg2022finite}. We refer to methods that develop Generalized Finite Differences (GFD) among the alternative approaches~\cite{song2020generalized,benito2003h}.
    As part of the T-Explainer's core, we are investigating alternatives to refine the $\mathbf{h}$ optimization process while maintaining computational efficiency.

    We conducted experiments and comparisons using tabular data to create classification models for simplicity and because it is a typical setup for machine learning problems in academia and industry~\cite{borisov2022deep}. 
    We highlight that our experimental setup followed the evaluation practices in the recent literature~\cite{baldi2014searching,tan2023considerations}, focusing on Neural Networks as black boxes and including datasets where leading XAI approaches have stability and applicability difficulties.
    However, we designed the T-Explainer considering more complex data representations, such as 2D images, videos, or semantic segmentation.
    We are extending our approach to include simplification masking approaches, a strategy widely used by many XAI methods~\cite{lime_ribeiro2016should,shap_lundberg2017unified}.
    
    The T-Explainer framework provides other metrics to evaluate feature importance in addition to those included in this publication, released as a Python module. We are actively extending this module to encompass more metrics and ensure their robustness, efficiency, and flexibility.
    A comprehensive evaluation of an explanation method requires quantitative~\cite{bodria2021benchmarking} and qualitative~\cite{shap_weerts2019human} assessments. 
    We prioritized quantitative metrics in this work, particularly those for evaluating consistency, continuity, and correctness~\cite{nauta2023from}.
    We are committed to comprehensively validating the T-Explainer using metrics beyond those discussed here.
    
    When carefully developed and applied, explainability contributes by adding new trustworthy perspectives to the broad horizon of Machine Learning, enriching the next generation of transparent {\AI} applications~\cite{lapuschkin2019unmasking}.
    As an evolving project, T-Explainer will continue to be expanded and updated, incorporating additional functionalities and refinements. Our goal is to provide a complete XAI suite in Python based on implementations of T-Explainer and its functionalities in a well-documented and user-friendly package.

%=============================================================================
\section{Conclusion}
\label{conclusion}
    In this paper, we present the T-Explainer, a Taylor expansion-based XAI technique. It is a deterministic local feature attribution method built on a solid mathematical foundation that grants it relevant properties. %such as local accuracy, missingness, and consistency. 
    Moreover, T-Explainer is designed to be a model-agnostic method capable of generating explanations for a wide range of machine learning models, as it relies only on the models' outputs without accounting for their internal structure or partial results. The experiments and comparisons demonstrate that the T-Explainer is stable and accurate compared to model-specific techniques, thus making it a valuable explanation alternative.

%=============================================================================
\begin{ack}
    This work was partially supported by the Coordena{\c{c}}{\~a}o de Aperfei{\c{c}}oamento de Pessoal de N{\'i}vel Superior--Brasil (CAPES)--Finance Code 001, by CNPq Grant 307184/2021-8, and by the São Paulo Research Foundation (FAPESP) Grant 2022/09091-8.
    The views expressed in this material are the authors' responsibility and do not necessarily reflect the views of the funding agencies.
\end{ack}

%=============================================================================
\small
\bibliography{references}

%=============================================================================
\newpage
\normalsize
\section{Appendix}
\label{appendix}
    Our $\mathbf{h}$ optimizer was designed to find a reasonable estimate of the displacement parameter $\mathbf{h}$ when running the centered finite difference method ($FD$) to compute the partial derivatives required by the T-Explainer. 
        The optimizer is based on a binary search that minimizes a cost function of {mean squared error} (MSE).
        The optimum $\mathbf{h}$ should be a sufficiently small value but not too small to avoid round-off errors or singularity cases and should not be too large to avoid truncation errors.

        Finite differences are well-established approaches to approximate differential equations, replacing the derivatives with discrete approximations (such transformation results in computationally feasible systems of equations)~\cite{leveque2007finite}.
        
        The algorithm~\ref{alg:h_opt} provides high-level details about the optimization process to estimate the parameter $\mathbf{h}$.
        If $\mathbf{h}$ leads to a singular or rank-deficient Jacobian matrix, the final ${\nabla}f_{\mathbf{x}}$ will lead the T-Explainer to attribute misleading null values to potentially important features. 
        Then, $\theta$ is a function that keeps the numerical stability by checking if the gradient of $f$ comes from a non-singular (or full-rank) matrix.
        
        \begin{algorithm}[H]
        \vspace{-0.1cm}
        \caption{The $\mathbf{h}$ optimizer. The input parameter $max\_itr$ defines the maximum number of iterations.}
        \label{alg:h_opt}
        \flushleft
        \textbf{Input:} $f$, $\mathbf{x}$, $h_{min}$, $h_{max}$, $max\_itr$ \\
        \textbf{Parameter:} $C$, a constant value independent of $\mathbf{h}$~\cite{leveque2007finite} \\
        \textbf{Output:} The optimum $\mathbf{h}$ and estimated ${\nabla}f_{\mathbf{x}}$
        \begin{algorithmic}[1] %[1] enables line numbers
            \STATE Let $h_{left} \gets h_{min}$, $h_{right} \gets h_{max}$, $prev\_cost \gets 1$
            \STATE $in\_loop \gets \text{True}$, $itr \gets 0$
            \WHILE{$in\_loop$ is \text{True}}
                \STATE $itr \gets itr + 1$
                \STATE $f_{\mathbf{x}} \gets f({\mathbf{x}})$
                \STATE $\mathbf{h} \gets (h_{left}+h_{right})/2$
                \STATE $\epsilon_{\mathbf{x}} \gets C*(h_{left})^2$
                \STATE ${\nabla}f_{\mathbf{x}} \gets FD(f, \mathbf{x}, \mathbf{h})$
                \STATE $\tilde{f}_{\mathbf{x}} \gets f(\mathbf{x} + \mathbf{h}) - {\nabla}f_{\mathbf{x}} \cdot \mathbf{h}$
                \STATE $curr\_cost \gets \mathcal{L}(f_{\mathbf{x}},\tilde{f}_{\mathbf{x}})$
                
                \IF {$curr\_cost > \epsilon_{\mathbf{x}}$}
                    \STATE $h_{right} \gets \mathbf{h}$
                \ELSE
                    \IF {$\theta({\nabla}f_{\mathbf{x}})$ is \text{True}}
                        \STATE $h_{left} \gets \mathbf{h}$
                    \ELSIF {$prev\_cost$ and $|curr\_cost - prev\_cost| < \epsilon_{\mathbf{x}}$}
                        \STATE $in\_loop \gets \text{False}$
                    \ELSE
                        \STATE $h_{left} \gets \mathbf{h}$
                    \ENDIF
                \ENDIF
                \STATE $prev\_cost \gets curr\_cost$
                \IF {$itr > max\_itr$}
                    \STATE $in\_loop \gets \text{False}$
                \ENDIF
            \ENDWHILE
            \STATE \textbf{return} $\mathbf{h}$ and ${\nabla}f_{\mathbf{x}}$
        \end{algorithmic}
        \end{algorithm}

%=============================================================================
\end{document}